\documentclass[12pt]{article}
\usepackage[figuresright]{rotating}
\usepackage{amsthm}
\usepackage{amssymb}
\usepackage{amsmath}
\usepackage{amsfonts}
\usepackage{array}
\usepackage{algorithm}
\usepackage{algorithmic}
\usepackage{subcaption}
\usepackage{multirow}
\usepackage{float}
\usepackage{caption}
\usepackage{graphicx}
\usepackage{booktabs}
\usepackage{enumitem}
\usepackage [colorlinks, linkcolor=blue, anchorcolor=blue, citecolor=blue, urlcolor=blue] {hyperref}

\newtheorem{pf}{Proof}
\newtheorem{thm}{Theorem}

\newtheorem{remark}{Remark}
\newtheorem{proposition}{Proposition}
\usepackage[dvipsnames]{xcolor}
\usepackage{bm}
\hypersetup{colorlinks,urlcolor=blue}
\usepackage{svg}
\usepackage{adjustbox}

\usepackage[numbers,sort&compress]{natbib}
\title{A privacy-preserving distributed credible evidence fusion algorithm for collective decision-making}
\author{Chaoxiong Ma \thanks{chaoxiongma@mail.nwpu.edu.cn} \and Yan Liang \thanks{liangyan@nwpu.edu.cn} \and Xinyu Yang \thanks{yang17691231609@163.com} \and Han Wu \thanks{wuhan@mail.nwpu.edu.cn} \and Huixia Zhang \thanks{zhanghuixia@mail.nwpu.edu.cn}}

\newcolumntype{R}[2]{%
    >{\adjustbox{angle=#1,lap=\width-(#2)}\bgroup}%
    l%
    <{\egroup}%
}
\date{} % 留空日期字段

\begin{document}

\def\spacingset#1{\renewcommand{\baselinestretch}%
{#1}\small\normalsize} \spacingset{1}

\maketitle
The theory of evidence reasoning has been applied to collective decision-making in recent years. However, existing distributed evidence fusion methods lead to participants' preference leakage and fusion failures as they directly exchange raw evidence and do not assess evidence credibility like centralized credible evidence fusion (CCEF) does. To do so, a privacy-preserving distributed credible evidence fusion method with three-level consensus (PCEF) is proposed in this paper. In evidence difference measure (EDM) neighbor consensus, an evidence-free equivalent expression of EDM among neighbored agents is derived with the shared dot product protocol for pignistic probability and the identical judgment of two events with maximal subjective probabilities, so that evidence privacy is guaranteed due to such irreversible evidence transformation. In EDM network consensus, the non-neighbored EDMs are inferred and neighbored EDMs reach uniformity via interaction between linear average consensus (LAC) and low-rank matrix completion with rank adaptation to guarantee EDM consensus convergence and no solution of inferring raw evidence in numerical iteration style. In fusion network consensus, a privacy-preserving LAC with a self-cancelling differential privacy term is proposed, where each agent adds its randomness to the sharing content and step-by-step cancels such randomness in consensus iterations. Besides, the sufficient condition of the convergence to the CCEF is explored, and it is proven that raw evidence is impossibly inferred in such an iterative consensus. The simulations show that PCEF is close to CCEF both in credibility and fusion results and obtains higher decision accuracy with less time-comsuming than existing methods.

\textbf{Keyword:} Credibility calculation, Distributed fusion, Evidence reasoning, Network consensus, Privacy-preserving, Collective decision-making.
%\begin{keyword}
%	Credibility fusion, Evidence fusion, Cyberattacks, Belief function, Distributed system
%\end{keyword}

\section{Introduction}
\label{sec:introduction}
In recent years, the development of sensors and ad hoc network techniques has promoted research on distributed inference \cite{hao2022distributed, de2019distributed}. As an important branch of distributed inference, collective decision-making \cite{kayaalp2023arithmetic} asks decentralized agents to collaboratively infer categorical variables of the global environment based on their local observations \cite{cai2023consensus, godoy2021grid}. For example, based on multi-view 3D data collected by sensors mounted on vehicles or infrastructure, road target categories are detected in a fusion center and then delivered to vehicles to cope with the problem of occluded vehicle vision \cite{arnold2020cooperative}. A vehicular cooperative perception approach including communication network repair and a vehicle-to-vehicle (V2V) attention module is designed in \cite{li2023learning} to reduce the negative impact of lossy communication on the perception task. In these distributed collaborative systems, low-accuracy sensors are widely used to reduce hardware costs, which leads to large measurement uncertainties and makes collective decision-making difficult.

As a common mathematical tool for representing and dealing with such uncertainty, the theory of evidence reasoning (ER) \cite{liu2022multisource} has widely applied to decision-level information fusion such as cyber-attack detection \cite{beechey2023evidential,uflaz2024quantifying}, risk analysis \cite{aydin2023analysis,lu2022riskassessment}, multi-criteria decision making \cite{sheng2024novel,liu2022multiattribute,hua2022consensus}, social learning \cite{liu2023asinusoidal,liu2023imprecise}, clustering and classification \cite{huang2023higher,zhang2023new,gan2023safe,gong2024self}, and so on. It therefore naturally draws the attention of researchers who focus on distributed evidence fusion. In \cite{lima2022evidential}, the belief function is used to estimate the confidence level of the output information from the collaborative perception participants. In \cite{bartashevich2021multi}, the distributed agents transform local explorations into pieces of evidence to perceive the dominant color of a closed square environment by fusing evidence with eight evidence fusion rules with a designed three-step positive feedback iterative modulation-driven mechanism. In \cite{zoghby_2014_evidential}, the ER is used to recognize vehicle types and construct dynamic maps to extend the perception of road vehicles. All these works are just simple applications of traditional evidence fusion methods to distributed systems.

In fact, the landmark work for distributed evidence fusion is provided by \cite{kanjanatarakul2017distributed}, in which the basic fusion rule of ER, i.e., Dempster's rule (DR), is naturally extended to the distributed systems for the first time. This work directly fuses the evidence of all agents based on the linear average consensus (LAC) algorithm and the commonality function representation of the evidence. It is proven to be equivalent to centralized DR in terms of fusion results and well adapted to both synchronous and asynchronous LAC mechanisms. However, the work does not address the two important issues of the distributed evidence fusion. On the one hand, the DR often produces counterintuitive results when fusing highly conflicting evidence \cite{shafer1976amathematical,abellan2021combination} and fails to cope with the cyclic propagation of evidence in networks gracefully. On the other hand, the direct sharing of raw evidence leads to participants' preference leakage.

To the best of our knowledge, there are two types of available solutions for counterintuitive issue. The first type designs alternative reasoning rules to reassign conflicting terms in various optimization criteria \cite{smets1990the,dubois1988representation,yager1987on}. In \cite{ducourthial2012self}, DR and cautious rule (CR) are mixed to fuse neighborhood and non-neighborhood evidence, respectively. This strategy is verified to be self-stabilizing and solves the data insect triggered by cyclic propagation, but the idempotent of CR also leads to the fusion result not changing when multiple sources hold the same evidence. Although \cite{guyard2018study} designed three combinatorial layouts containing DR and CR to solve the data insect problem, the introduced alternative rules still carry inherent flaws. On the one hand, alternative fusion rules often lose commutativity and associativity \cite{xiao2019multi}, which are favorable for evidence fusion in a distributed manner. On the other hand, they are often invalid in dealing with high conflicts resulting from sensor failures. 

Therefore, another type of solution, so-called credible evidence fusion (CEF), is more favored by researchers in centralized evidence fusion. The CEF retains DR while pre-processing evidence to reduce conflicts according to evidence credibility \cite{murphy2000combining,yong2004combining}, thus being an extension of Bayesian theory and consistent with human reasoning. Available distributed evidence fusion assigns credibility to information sources based on a priori external factors. For example, in road icing monitoring scenarios, the credibility of evidence is considered to be positively correlated with its propagation distance \cite{guyard2018study}. In critical infrastructure operational status monitoring, the quality of source historical evidence and the importance of network links are adopted as references for judging source credibility \cite{pietro2015situational}. In \cite{zoghby_2014_evidential}, a fixed discount factor is imposed on the evidence to avoid high conflict leading to non-convergence of fusion. Although this a priori information-dependent subjective credibility partially solves the high conflict problem, it suffers from poor timeliness, poor access to a priori information, and poor adaptability to real-time data, which leads to potentially low fusion accuracy. Recently, a class of distributed evidence fusion strategies based on anomalous detection can be regarded as a binarized evidence credibility assessment process. Based on the random sample consensus (RANSAC) algorithm, distributed evidence consensus algorithms for DR and CR are proposed in \cite{denux2021distributed}. Also based on the idea of outlier detection, \cite{zhao2023information} employs the connectivity-based outlier factor (COF) method to remove disturbed evidence. These outlier-detection-based methods require adjusting algorithm parameters to balance false positive and false negative rates, where a high false positive rate leads to the detachment of normal information sources, affecting fusion accuracy, and a high false negative rate treats interfering evidence as normal, leading to fusion failure. Therefore, it is still imperative to investigate methods for credibility calculation in distributed systems. Traditional centralized CEF (CCEF) assesses credibility based on the evidence difference measure matrix (EDMM), which is based on two-by-two comparisons between pieces of evidence and has better adaptability \cite{yong2004combining,liu2011combination,xiao2020anew}. However, how to construct EDMM and assess evidence credibility in a distributed system is still unresolved.

The privacy protection of raw data is also an important topic for collective decision-making \cite{xiong2022toward,alemany2022review}. Although evidence only expresses the occurrence probabilities of events in the publicly available framework of discernment (FoD), they still reflect sideways certain preferences of agent data, such as the business status of enterprises in economic surveys, the content of voters' votes in social elections, and the details of the combat unit's missions in military confrontations. Therefore, it is necessary to protect the privacy of agent evidence. Most of the existing distributed evidence fusion solutions require that evidence from neighbors be available, which definitely conflicts with privacy preservation. In other words, distributed CEF with privacy preservation remains an open and interesting problem.

In light of the aforementioned analysis, this paper aims to develop a privacy-preserving distributed credible evidence fusion method for collective decision-making with the premise of protecting agents' raw evidence data. The main contributions include:
\begin{enumerate}[label=\arabic*)]
	\setlength{\itemsep}{0pt}
	\setlength{\parsep}{0pt}
	\setlength{\parskip}{0pt}
	\item The privacy-preserving distributed credible evidence fusion (DCEF) problem is proposed for the need to deal with information uncertainty in collective decision-making, in which agents are required to fuse their evidence in a distributed manner and prevent their raw evidence from being known or inferred by others. The distributed fusion result is expected to converge to CCEF, which has shown good engineering adaptability.
	\item A fully privacy-preserving distributed credible evidence fusion algorithm (PCEF) is designed. It has a three-level consensus: the evidence difference measure (EDM) neighbor consensus, the EDM network consensus, and the fusion network consensus.
	\item Considering that the unavailability of raw evidence from other agents makes EDMM construction difficult, a privacy-preserving distributed construction strategy for EDMM covering EDM neighbor consensus and EDM network consensus is given, where the EDM neighbor consensus allows neighbors to calculate EDM between pieces of raw evidence without disclosure. The EDM network consensus locates the locally constructed EDMM to each agent and recovers the EDM between non-neighboring agents' pieces of raw evidence by using a low-rank matrix complementation technique that has rank-adaptive capability.
	\item A distributed fusion strategy based on differential privacy and credibility compensation is developed to prevent agents' raw evidence from being inferred during the fusion network consensus, and its capability for privacy preservation is discussed.
\end{enumerate}

This paper is organized as follows: Section \ref{sec: Problem formulation} analyzes the challenges of DCEF with privacy-preserving; Section \ref{sec: The PCEF algorithm} presents the details of the PCEF; Section \ref{sec:Simulation} simulates and verifies the effectiveness of PCEF; and the work is finally summarized.
\section{Problem formulation}
\label{sec: Problem formulation}
\begin{figure*}[!h]
	\centering
	\includegraphics[width=0.75\linewidth]{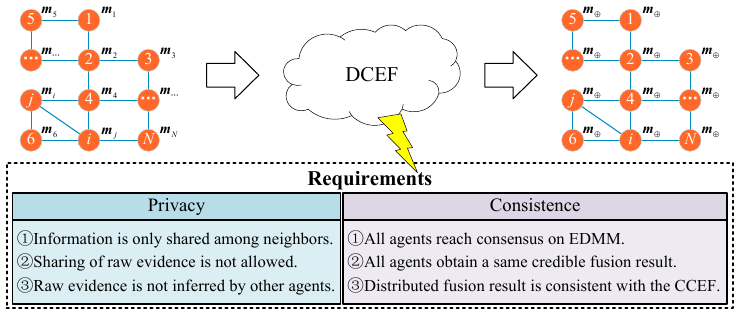}
	\caption{The requirements of DCEF with privacy preservation of raw evidence.}
	\label{fig1:Fig_1_Profor_challenge}
\end{figure*}

As shown in Fig.\ref{fig1:Fig_1_Profor_challenge}, $N$ distributed agents constitute a distributed system that is modeled as a strongly connected undirected graph. ${\cal G} = \left({\cal V},\varepsilon \right)$, where ${\cal V} = \{ 1, 2, \cdots, N \}$ and $\varepsilon \subset \cal V \times \cal V$ are the set of agents and edges, respectively. Two agents connected by an edge $e=(i,j) \in \varepsilon$ are called neighbors. The set of neighbors of Agent $i$ is expressed as ${\cal N}_i = \{ j | \left(i,j\right)  \in \varepsilon  \}$. Let ${\cal A}=[a_{ij}] \in {\mathbb{R}^{N \times N}}$ and ${\cal A}^i=[a^i_{ij}] \in {\mathbb{R}^{N \times N}}$ be the adjacency matrix of $\cal G$ and the local adjacency matrix of Agent $i$, respectively, with $a_{ij} = a^i_{ij} = 1$ for $(i,j) \in \varepsilon$ and $a_{ij} = a^i_{ij} = 0$ for otherwise. 

The $N$ agents perform a collective decision-making task in which all complete and mutually exclusive potential decision results form a finite set, FoD~\cite{liu2020determine}, written as $\Omega = \{ {\hat A}_1,{\hat A}_2, \cdots,{\hat A}_n \}$. Here, ${\hat A}_i$ is a potential decision result. During the decision-making process, the agent's information is uniformly characterized in the FoD space to cope with the heterogeneity and uncertainty of the source data. In the beginning, each of the $N$ agents holds a piece of evidence, also called basic belief assignment (BBA) or mass function, defined on the FoD. Note that Agent $i$'s raw evidence $\boldsymbol{m}_i$ is a mapping ${m_i}:{2^\Omega} \to \left[ {0,1} \right]$ such that:
% The evidence ${\boldsymbol{m}_i} = {\left[{m_i}\left( \emptyset \right),{m_i}\left( {{{\hat A}_1}} \right),{m_i}\left( {{{\hat A}_2}} \right), \cdots,{m_i}\left( \Omega \right)\right]^T}$ is called BBA or mass function, mapping ${m_i}:{2^\Omega } \to \left[ {0,1} \right]$ from the power set ${2^\Omega }{\text{ = }}\{ {\emptyset,{A_1},{A_2}, \cdots,{A_r}} \}$ with ${A_i} \subseteq \Omega$ and $r = {2^n} - 1$ to $\left[ {0,1} \right]$, and it satisfies the following constraint:The evidence $\boldsymbol{m}_i$, called BBA or mass function, is a mapping ${m_i}:{2^\Omega } \to \left[ {0,1} \right]$ such that:
\begin{equation}
\sum\nolimits_{A \subseteq \Omega } m_i\left( A \right) = 1
\label{eq1:EvidenceDefinition}
\end{equation}
and, in general, ${m_i}\left( \emptyset \right) = 0$.

\begin{figure*}[!h]
	\centering
	\includegraphics[width=0.7\linewidth]{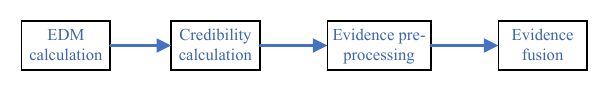}
	\caption{Overview of CCEF.}
	\label{fig3:Fig3_CCEF_Algorithm}
\end{figure*}
The communication constraints of agents and the task goals for the collective decision-making task are shown in Fig.\ref{fig1:Fig_1_Profor_challenge}. Agents are required to collaboratively fuse all pieces of evidence. During this process, data sharing is limited to authenticated neighboring agents to prevent exposure of agent preferences. In other words, the raw evidence of an agent is not shared with or inferred by any other agents. As there is no fusion center, agents have to exchange and update data iteratively to reach consensus. The term ``consensus" encompasses three aspects: first, the basis for credibility calculation, i.e., the EDMM, is consistent; second, all agents obtain the same fusion result; and third, the fusion result accurately converges to the CCEF. As shown in Fig.\ref{fig3:Fig3_CCEF_Algorithm}, the CCEF procedure is outlined below at first:
\begin{enumerate}[label=\arabic*)]
	\item \textbf{EDM calculation}: Compute EDMs among $N$ pieces of evidence. Here, a commonly used EDM, $Dismp$~\cite{liu2011combination}, is shown:
	\begin{equation}
	d_{ij} = DismP( \boldsymbol{m}_i,\boldsymbol{m}_j ) \triangleq \frac{{DistP( \boldsymbol{m}_i,\boldsymbol{m}_j ) + ConfP( \boldsymbol{m}_i,\boldsymbol{m}_j )}}{{1 + DistP( \boldsymbol{m}_i,\boldsymbol{m}_j )ConfP( \boldsymbol{m}_i,\boldsymbol{m}_j)}}
	\label{eq3:DismPDefinition}
	\end{equation}
	where $d_{ij}$ is the EDM between $\boldsymbol{m}_i$ and $\boldsymbol{m}_j$, and
	\begin{equation}
	\begin{aligned}
	DistP\left( {\boldsymbol{m}_i,\boldsymbol{m}_j} \right) = 
	\frac{1}{2}{\left( {\sum\nolimits_{\scriptstyle {{\hat A}_k} \in \Omega , \atop 	\scriptstyle \left| {{{\hat A}_k}} \right| = 1} {{{\left| {BetP_i({\hat A}_k) - BetP_j( {\hat A}_k )} \right|}^\eta }} } \right)^{\frac{1}{\eta }}}
	\end{aligned}
	\label{eq4:DistPDefinition}
	\end{equation}
	\begin{equation}
	\begin{aligned}
	ConfP\left(\boldsymbol{m}_i,\boldsymbol{m}_j \right) &= 
	\begin{cases}
	0&, \text{if }X_{\max }^{\boldsymbol{m}_i} \cap X_{\max }^{\boldsymbol{m}_j} \ne \emptyset \\
	BetP_i\left( {X_{\max }^{\boldsymbol{m}_i}} \right)BetP_j\left( {X_{\max }^{\boldsymbol{m}_j}} \right)&, \text{otherwise.}
	\end{cases}\\
	s.t.\quad X_{\max }^{{\boldsymbol{m}_l}} &= \arg \mathop {\max }\limits_{{{\hat A}_k} \in \Omega } {BetP_{{\boldsymbol{m}_l}}}({{\hat A}_k}),l = i,j
	\end{aligned}
	\label{eq5:ConfPDefinition}
	\end{equation}
	where $\eta = 2$; $DistP$ and $ConfP$ are respectively the probabilistic-based distance and the conflict coefficient. They are calculated by Pignistic probability, i.e., $\boldsymbol{BetP}_i=[{BetP_i ( {\hat A}_1 ), BetP_i( {\hat A}_2 ),\cdots, BetP_i( {\hat A}_n )} ]^T$:
	\begin{equation}
	BetP_i\left( {{{\hat A}_k}} \right) = \mathop \sum \limits_{{B} \subset {2^\Omega },{{\hat A}_k} \subseteq B} \frac{1}{{|B|}}{m_i}\left( B \right)
	\label{eq6:BetPDefinition}
	\end{equation}
	where $|B|$ is the cardinality of $B$.
	\item \textbf{Credibility calculation}: Compute the credibility $Cred_i$ of $\boldsymbol{m}_i$:
	\begin{equation}
	Cred_{i} = \frac{{\mathop {\min }\limits_{1 \le j \le N} \sum\nolimits_{k = 1,k \ne j}^N {{d_{jk}}} }}{{\sum\nolimits_{k = 1,k \ne i}^N {{d_{ik}}} }}
	\label{eq7:CredibilityDefinition}
	\end{equation}
	\item \textbf{Evidence pre-processing}: Furthermore, raw evidence is discounted in a crediability-dependent style to reduce conflicting~\cite{zhang2022information}:
	\begin{equation}
	\begin{cases}
	{{m'}_i}(A) = Cred_i \cdot {m_i}(A)&, \text{if }A \ne \Omega \\
	{{m'}_i}(\Omega ) = 1 - \sum\limits_{A \in {2^\Omega },{}A \ne \Omega } {{{m'}_i}(A)}&, \text{otherwise.}
	\end{cases}
	\label{eq8:EvidenceDiscount}
	\end{equation}
	\item \textbf{Evidence fusion}: Fuse all pre-processed evidence to obtain the fusion result $\boldsymbol{m}_{\oplus}$:
	\begin{equation}
	\boldsymbol{m}_{\oplus} = \boldsymbol{m}'_1 \oplus \boldsymbol{m}'_2 \oplus \cdots \oplus \boldsymbol{m}'_N
	\label{eq9:DRforNEvidence}
	\end{equation}
	where $\oplus$ is the DR operator. Since DR follows the associativity, $\oplus$ is shown with $\boldsymbol{m}_i$ and $\boldsymbol{m}_j$ as examples:
	\begin{equation}
	\left( {{m_i} \oplus {m_j}} \right)\left( A \right) = 
	\begin{cases}
	0 &, \text{if } A = \emptyset \\
	\frac{{\sum\limits_{B \cap C = A} {{m_i}\left( B \right){m_j}\left( C \right)} }}{{1 - {\sum _{B \cap C = \emptyset }}{m_i}\left( B \right){m_j}\left( C \right)}} &, \text{if } A \in 2^\Omega \backslash \{\emptyset\}
	\end{cases}
	\label{eq10:DRCentralized}
	\end{equation}
\end{enumerate}
%\textbf{The discussion about $\eta = 2$. }

As for distributed evidence fusion, available studies reach all-agent consensus on fusion result via LAC~\cite{denux2021distributed,kanjanatarakul2017distributed}:
\begin{equation}
\begin{aligned}
\boldsymbol{x}_i^{\boldsymbol{\omega}} \left( {t + 1} \right) &= \boldsymbol{x}_i^{\boldsymbol{\omega}} \left( t \right) + \sum\limits_{j \in {{\cal N}_i}}^N {{c_{ij}}\left( {\boldsymbol{x}_j^{\boldsymbol{\omega}} \left( t \right) - \boldsymbol{x}_i^{\boldsymbol{\omega}} \left( t \right)} \right)} \\
\boldsymbol{x}_i^{\boldsymbol{\omega}} \left( {0} \right) &= \boldsymbol{\omega}_i
\end{aligned}
\label{eq30:EvidenceFusion_LAC}
\end{equation}
where $\boldsymbol{x}_i^{\boldsymbol{\omega}}\left( t \right)$ represents the state of Agent $i$ at iteration step $t$; $c_{ij}$ is the Metropolis-Hastings weight~\cite{xiao2005ascheme}:
\begin{equation}
c_{ij} = \begin{cases}
\frac{1}{\max\{\left| \mathcal{N}_i \right|,\left| \mathcal{N}_j \right|\}+1} &,  \text{if } j\in \mathcal{N}_i \text{ and }j\ne i \\ 
1- \sum\nolimits_{j \in {\cal N}_i} \frac{1}{\max\{\left| \mathcal{N}_i \right|,\left| \mathcal{N}_j \right|\}+1} &,  \text{if } j = i \\
0&,\text{otherwise}
\end{cases}
\label{eq15:LinearConsensusWeight}
\end{equation}
where $\left| {\mathcal{N}}_i \right|$ is the cardinality of ${\mathcal{N}}_i$. The $\boldsymbol{\omega}_i$ is the weight assignment of $\boldsymbol{m}_i$~\cite{smets1995canonical}. It is another representation of evidence. For $\forall A \in {2^\Omega }\backslash \left\{ {\emptyset ,\Omega } \right\}$, $\omega_i \left( A \right)$ is calculated by:
\begin{equation}
\omega_i \left( A \right) = \sum\limits_{B \supseteq A} {{{\left( { - 1} \right)}^{\left| B \right| - \left| A \right|}}\ln Q_i\left( B \right)}
\label{eq31:CommonalityFunction2WeightAssignment}
\end{equation}
where $Q_i\left( A \right):{2^\Omega } \to \left[ {0,1} \right]$ is the so-called commonality function that is converted by the mass function:
\begin{equation}
Q_i\left( A \right) = \sum\limits_{B \supseteq A} {m_i\left( B \right)}
\label{eq32:CommonalityFunction}
\end{equation}
The Eq.(\ref{eq30:EvidenceFusion_LAC}) will lead to the fact that the states of all agents converge to $\sum\nolimits_{i = 1}^N {{\boldsymbol{\omega}_i}} /N$ when ${\cal G}$ is a strongly connected graph. Considering that the fusion of mass functions with DR is equivalent to the addition of weight assignments, i.e., the weight assignment corresponding to the $\boldsymbol{m}_i \oplus \boldsymbol{m}_j$ is $\boldsymbol{\omega }_i + \boldsymbol{\omega}_j$, so $\sum\nolimits_{i = 1}^N {{\boldsymbol{\omega}_i}} /N$ is exactly $1/N$ of the fusion result of all pieces of evidence with DR. Let $\boldsymbol{\omega}_\oplus \triangleq \sum\nolimits_{i = 1}^N {{\boldsymbol{\omega}_i}}$ be the weight assignment of $\boldsymbol{m}_\oplus$, then:
\begin{equation}
\boldsymbol{m}_\oplus = A_1^{\boldsymbol{\omega}_\oplus} \oplus A_2^{\boldsymbol{\omega}_\oplus} \oplus \cdots \oplus A_{r}^{\boldsymbol{\omega}_\oplus} \triangleq \mathop \oplus \limits_{k = 1}^{r} A_k^{\boldsymbol{\omega}_\oplus}
\label{eq33:WeightAssignment2MassFunction}
\end{equation}
where $A_k^{\boldsymbol{\omega}_\oplus}$ is the simple mass function~\cite{denux2021distributed} corresponding to $\omega_\oplus ({A_k})$, and $r = 2^n - 2$.

According to the analysis of CCEF and existing distributed evidence fusion above, there are several challenges to realizing the collective decision-making task shown in Fig.\ref{fig1:Fig_1_Profor_challenge}: 
\begin{enumerate}[label=\arabic*)]
	\setlength{\itemsep}{0pt}
	\setlength{\parsep}{0pt}
	\setlength{\parskip}{0pt}
	\item In EDM calculation, Eq.(\ref{eq3:DismPDefinition}) asks agents to collect a pair of raw evidence simultaneously, which is hindered by the privacy requirements, i.e., the forbiddings of raw evidence sharing and non-neighbors information sharing.
	%Calculation of EDM is prerequisite for credibility assessment, which askes agents to collecte pair of raw evidence simultaneously. On the one hand, revealing raw evidence to others contradicts privacy protection requirements. On the other hand, the lack of information sharing among non-neighboring agents hinders EDM calculation.
	\item In credibility calculation, all EDMs among pieces of evidence are involved in Eq.(\ref{eq7:CredibilityDefinition}). Therefore, it is essential to ensure that all agents possess identical EDMs.
	%As shown in Eq.(\ref{eq7:CredibilityDefinition}), credibility calculation involves all EDMs. Therefore, it is essential to ensure that all agents access to and possess identical EDMs.
	\item In evidence fusion, uniform EDMs mean that evidence credibilities are publicly accessible to all agents, and hence the pre-processed evidence and raw evidence can be derived from each other according to Eq.(\ref{eq8:EvidenceDiscount}). In other words, taking the pre-processed evidence as the initial state of Eq.(\ref{eq30:EvidenceFusion_LAC}) is not privacy-preserving.
	%	If agents obtain identical EDMs, the evidence credibilities are public, and hence the discounted evidence and raw evidence can be derived by each other according to Eq.(\ref{eq8:EvidenceDiscount}). In other words, taking the pre-processed evidence as the initial state of Eq.(\ref{eq30:EvidenceFusion_LAC}) is not privacy-preserving. 
	%Hence, it is urgent to explore privacy-preserving sharing and consensus mechanism.
\end{enumerate}

%\begin{equation}
%D = \left[ d_{ij} \right] \in {\mathbb{R}^{N \times N}}
%\label{eq2:EDMMDefinition}
%\end{equation}
% works in a completely private distributed network to achieve credible evidence fusion

\section{The PCEF algorithm}
\label{sec: The PCEF algorithm}

\begin{figure}[!h]
	\centering
	\includegraphics[width=0.75\linewidth]{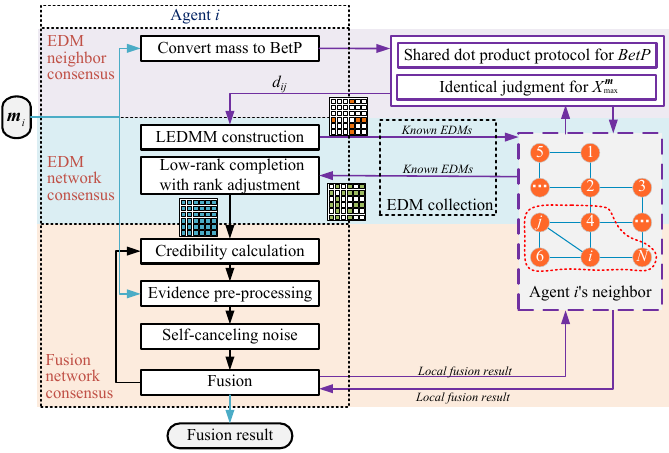}
	\caption{The flowchart of PCEF}
	\label{fig2:Fig2_PCEF_Algorithm}
\end{figure}

To address the three challenges of CEF in a fully privacy-preserving distributed network, a method named PCEF that includes EDM neighbor consensus, EDM network consensus, and fusion network consensus is proposed in this section, as shown in Fig.\ref{fig2:Fig2_PCEF_Algorithm}. In EDM neighbor consensus, neighboring agents compute the EDM between their raw evidence without knowledge of each other's raw evidence and $BetP$. In EDM network consensus, the EDM of non-neighboring agents' raw evidence is estimated based on the correlation between the rows of the EDMM and the already-obtained EDMs between neighbors' raw evidence. The result of this consensus is that each agent holds a complete, all-agent-consistent EDMM, which will be used for credibility computation. In fusion network consensus, differential privacy and credibility compensation are introduced into Eq.(\ref{eq30:EvidenceFusion_LAC}) for a finite time to protect the privacy of raw evidence and lead all agents to achieve consensus on fusion.

\subsection{EDM neighbor consensus}
\label{subsec: EDM neighbored consensus}
Although the computation of $Dismp$ does not involve accurate information about $\boldsymbol{m}_i$ and $\boldsymbol{m}_j$, $\boldsymbol{BetP}$ still features the agent's preferences. Therefore, the privacy of $\boldsymbol{BetP}$ is also required to be protected during the computation of $Dismp$ besides the raw evidence. To this end, this section utilizes two secure multi-party computation protocols, namely the inner product computation of Pignistic probabilities and the identical judgment of two events with maximal subjective probabilities, to compute the EDM of neighboring agents' raw evidence while protecting the privacy of the agent's raw evidence and $\boldsymbol{BetP}$.

The privacy calculation of $DistP$ is presented next. According to Eq.(\ref{eq4:DistPDefinition}), $DistP$ can be written as:
\begin{equation}
\begin{array}{cl}
DistP\left( {{\boldsymbol{m}_i},{\boldsymbol{m}_j}} \right) 
&= \frac{1}{\sqrt 2}\left( \sum\limits_{ {\hat A}_k \in \Omega ,\left| {\hat A}_k \right| = 1} {{\left| BetP_i\left( {\hat A}_k \right) - BetP_j\left( {\hat A}_k \right) \right|}^2} \right)^{\frac{1}{2}}\\
&= \frac{1}{\sqrt 2}\left( \boldsymbol{BetP}_i - \boldsymbol{BetP}_j \right)^T \left( \boldsymbol{BetP}_i - \boldsymbol{BetP}_j \right)\\
&= \frac{1}{{\sqrt 2 }}\left( {\left\langle \boldsymbol{BetP}_i,\boldsymbol{BetP}_i \right\rangle + \left\langle \boldsymbol{BetP}_j,\boldsymbol{BetP}_j \right\rangle - 2\left\langle \boldsymbol{BetP}_i,\boldsymbol{BetP}_j \right\rangle } \right)
\end{array}
\label{eq10:DistPDotProDec}
\end{equation}
where $\left\langle \centerdot,\centerdot \right\rangle$ is the dot product operation. It is apparently simple and safe to compute $\langle \boldsymbol{BetP}_i, \boldsymbol{BetP}_i\rangle$ locally at Agent $i$ and send it to neighboring agents. But $\langle \boldsymbol{BetP}_i, \boldsymbol{BetP}_j\rangle$ cannot be computed, as it demands the agent knows both $\boldsymbol{BetP}_i$ and $\boldsymbol{BetP}_j$. Fortunately, the available privacy-preserving two-party shared dot product protocol~\cite{hu2020privacy,hu2017efficient} allows neighboring agents to obtain $\langle \boldsymbol{BetP}_i, \boldsymbol{BetP}_j\rangle$ without knowing anything about each other's $\boldsymbol{BetP}$. For more details about two-party shared dot product protocol implementation and its mechanism discussion, see~\cite{wang2009design}.
\begin{remark}
	Generally, the value of \(\eta\) is recommended to be chosen as a positive integer as small as possible~\cite{liu2011combination}. The Eq.(\ref{eq4:DistPDefinition}) takes \(\eta = 2\) to ensure the privacy-preserving computation of \(DistP\) while reducing computational complexity. When \(\eta = 1\), the \(DistP\) is too simple to design a calculation protocol that preserves \(BetP\).
\end{remark}

%$\forall \hat{A}_k \in \Omega$, neighbors obtain $a_k$ and $b_k$ based on the Oblivious Transfer, respectively. And the shares of $a_k$ and $b_k$ in $BetP_i\left( {\hat A}_k \right) - BetP_j\left( {\hat A}_k \right) $ are random. As a result, Agents $i$ and $j$, respectively, are in possession of $\sum\nolimits_{k = 1}^n a_k $ and $\sum\nolimits_{k = 1}^n b_k $, which are portions of $\langle \boldsymbol{BetP}_i, \boldsymbol{BetP}_j\rangle$. After that, neighbors encrypt $\sum\nolimits_{k = 1}^n a_k $ and $\sum\nolimits_{k = 1}^n b_k $ with Paillier's homomorphic encryption to calculate $\langle \boldsymbol{BetP}_i, \boldsymbol{BetP}_j\rangle$, because the encryption allows for the addition operation to be performed directly on the ciphertext while maintaining the sum, which ensures the inner product is correct and prevents the communication between the involved parties from being eavesdropped on or tampered with by an attacker. 

The $ConfP$ is the segmentation function of the maximal subjective probability of $\boldsymbol{BetP}$, being zero in $X_{\max}^{\boldsymbol{m}_i} = X_{\max}^{\boldsymbol{m}_j}$, or $BetP_i\left(X_{\max}^{\boldsymbol{m}_i}\right)BetP_j\left(X_{\max }^{\boldsymbol{m}_j}\right)$ for different. Hence, it is critical to determine if the serial numbers $\mathchar'26\mkern-10mu k_i$ and $\mathchar'26\mkern-10mu k_j$ of $X_{\max}^{\boldsymbol{m}_i}$ and $X_{\max}^{\boldsymbol{m}_j}$ are equal. In the millionaire problem~\cite{liu2022design,saxena2020continuous,he2013simple}, two rich men compare their assets without the help of a third party, where each richer tries to prevent his opponent from knowing his wealth, so that the solution is aptly utilized to compare whether $\mathchar'26\mkern-10mu k_i$ and $\mathchar'26\mkern-10mu k_j$ are equal while keeping the neighboring agents as far away from the exact numbered values as possible. If the numbers are equal, Agent $i$ learns about $\mathchar'26\mkern-10mu k_j$, and Agent $j$ learns about $\mathchar'26\mkern-10mu k_i$ without exchanging information about $BetP_i\left(X_{\max}^{\boldsymbol{m}_i}\right)$ and $BetP_j\left(X_{\max}^{\boldsymbol{m}_j}\right)$. If the numbers are not equal, Agent $i$ informs Agent $j$ about $BetP_i\left(X_{\max}^{\boldsymbol{m}_i}\right)$, and vice versa, but they do not exchange $\mathchar'26\mkern-10mu k_j$ and $\mathchar'26\mkern-10mu k_i$. This ensures the privacy of $\boldsymbol{BetP}$, as the serial number and maximal subjective probability are not disclosed simultaneously to the other agent.

The following Alg.\ref{alg1:LEDMMConstruction} shows how the neighboring agents calculate neighbored EDM.

\begin{algorithm}[!h]
	\caption{EDM neighbor consensus.}
	\label{alg1:LEDMMConstruction}
	\begin{algorithmic}[1]
		\REQUIRE Evidence $\boldsymbol{m}_i$ held by Agent $i$ and $\boldsymbol{m}_j$ held by Agent $j$.
		\ENSURE EDM $d_{ij}$. %%output
		\IF {$i = j$}
		\STATE $d_{ij} \leftarrow 0$.
		\ELSIF {$(i,j) \in \varepsilon$}
		\STATE Agent $i$ converts $\boldsymbol{m}_i$ to $\boldsymbol{BetP}_i$, Agent $j$ converts $\boldsymbol{m}_j$ to $\boldsymbol{BetP}_j$.
		\STATE Agent $i$ calculates and sends $\left\langle \boldsymbol{BetP}_i,\boldsymbol{BetP}_i \right\rangle$ to Agent $j$, Agent $j$ calculates and sends $\left\langle \boldsymbol{BetP}_j,\boldsymbol{BetP}_j \right\rangle$ to Agent $i$.
		\STATE Agents $i$ and $j$ compute $\left\langle {\boldsymbol{BetP}_i,\boldsymbol{BetP}_j} \right\rangle$ jointly via the privacy-preserving dot product protocol~\cite{wang2009design}.
		\STATE Agents $i$ and $j$ compute $DistP(\boldsymbol{BetP}_i,\boldsymbol{BetP}_j)$ with Eq.(\ref{eq10:DistPDotProDec}).
		\STATE Agent $i$ gets the serial number $\mathchar'26\mkern-10mu k_i$ of $X_{max}^{\boldsymbol{m}_i}$, Agent $j$ gets the serial number $\mathchar'26\mkern-10mu k_j$ of $X_{max}^{\boldsymbol{m}_j}$.
		\STATE Agents $i$ and $j$ compare $\mathchar'26\mkern-10mu k_i$ and $\mathchar'26\mkern-10mu k_j$ jointly according to the solution for Millionaire Problem~\cite{liu2022design}.
		\IF {$\mathchar'26\mkern-10mu k_i \ne \mathchar'26\mkern-10mu k_j$}
		\STATE Agent $i$ sends $BetP_i\left(X_{\max }^{\boldsymbol{m}_i}\right)$ to Agent $j$, and Agent $j$ sends $BetP_j\left(X_{\max }^{\boldsymbol{m}_j}\right)$ to Agent $i$.
		\STATE Agents $i$ and $j$ compute $ConfP(\boldsymbol{m}_i,\boldsymbol{m}_j) \leftarrow BetP_i\left(X_{\max }^{\boldsymbol{m}_i}\right) BetP_j\left(X_{\max }^{\boldsymbol{m}_j}\right)$.
		\ELSE
		\STATE Agents $i$ and $j$ compute $ConfP(\boldsymbol{m}_i,\boldsymbol{m}_j) \leftarrow 0$.
		\ENDIF \\
		Agents $i$ and $j$ compute $Dismp(\boldsymbol{m}_i,\boldsymbol{m}_j)$ according to Eq.(\ref{eq3:DismPDefinition}).
		\ENDIF
	\end{algorithmic}
\end{algorithm}

%The LEDMM constructed by the agent only fills the EDM between the agent's raw evidence and that of its neighbors, since the distributed network allows the sharing and transmitting of information only between neighboring agents. Without loss of generality, all the unknown/missing elements in LEDMM are here set to 0. Take the LEDMM $\bar{D}_i$ of Agent $i$ as an example:

%\begin{equation}
%	\begin{aligned}
%	\bar{D}_i = &
%	\begin{bmatrix}
%	 0  & d_{12} & \cdots & d_{1N}\\
%	 d_{21}& 0  & \cdots & d_{2N}\\
%	 \vdots& \vdots & \ddots & \vdots\\
%	 d_{N1}& d_{N2} & \cdots & 0
%	\end{bmatrix} \in {\mathbb{R}}^{N \times N} \\
%	 s.t.\quad d_{ij} = &\left\{ 
%	\begin{aligned}
%	Dismp({\boldsymbol{m}_i},{\boldsymbol{m}_j}),\quad &{\rm{if }}\quad j \in {\cal{N}}_{i} \\
%	0,\quad &{\rm{otherwise}}
%	\end{aligned}
%	\right.\\
%	\end{aligned}
%\label{eq11:LEDMMDefination}
%\end{equation}

\subsection{EDM network consensus}
\label{subsec: EDM network consensus}
The absence of information sharing between non-neighbors hinders the computation of non-neighboring EDMs. Even though many evidence sources suffer from interference and measurement noise, there are still only a few pieces of evidence that contradict the correct event completely, which is why credibility is based on the ``majority-minority" principle. In other words, the differences among the majority of pieces of evidence are limited. Therefore, the matrix formed by EDMs is often row-correlated and hence brings out a somewhat low-rank property. On the principle that simpler is more reliable, the idea of low-rank/sparse matrix processing has been widely used in image processing~\cite{qin2022low}, compressed sensing~\cite{tanner2023compressed}, and other fields. In this subsection, the low-rank matrix completion technology is used to estimate the non-neighbored EDMs.
\subsubsection{EDM collection}
\label{subsubsec:EDMCollection}
In EDM neighbor consensus, the neighbored EDMs obtained by Agent $i$ in Alg.\ref{alg1:LEDMMConstruction} can be represented as a matrix called Local EDMM (LEDMM): ${\bar{D}_i} \triangleq {P_{{\mathcal{A}}_i}}( D) = {{\mathcal{A}}_{i}}\odot D \in {\mathbb{R}}^{N \times N}$, where $\odot $ is the Hadamard product and $D \triangleq [d_{ij}] \in {\mathbb{R}}^{N \times N}$ is the EDMM. In ${\bar{D}_i}$, the unknown non-neighbored EDMs are set to be zeros, indicating an optimistic assumption that non-neighbor evidence is completely consistent. Since more known elements improve the accuracy of matrix completion, all neighboring EDMs are first backed up to each agent based on LAC:
% to obtain $P_{\mathcal{ A}}\left( D \right)$
\begin{equation}
\begin{aligned}
D_i\left( t + 1 \right) &= D_i\left( t \right) + \sum\nolimits_{j \in {{\cal N}_i}} {c_{ij}\left( D_j\left( t \right) - D_i\left( t \right) \right)}\\
D_i\left( 0 \right) &= \bar{D}_i
\end{aligned}
\label{eq14:LAC}
\end{equation}
where $D_i\left( t \right)$ is the EDMM with missing elements held by Agent $i$ at $t$-th iterative step, and $c_{ij}$ is determined by Eq.(\ref{eq15:LinearConsensusWeight}). As $\sum\nolimits_{i\in \mathcal{V}}{{{\mathcal{A}}_{i}}}=2\mathcal{A}$ is held for the fixed topological undirected graph, we have $\sum\nolimits_{i\in \mathcal{V}}{{\bar{D}_{i}}}=\sum\nolimits_{i\in \mathcal{V}}{P_{\mathcal{A}_i}( D)}=2 {P_{\mathcal{A}}}(D)$. According to the LAC, the state $D_i\left( t \right)$ of Agent $i$ converges to:
\begin{equation}
\underset{t\to \infty }{\mathop{\lim }}\,{D_i}\left( t \right)=\frac{1}{N}\sum\limits_{i=1}^{N}{{D_{i}}\left( 0 \right)}=\frac{2}{N}{{P}_{\mathcal{A}}}\left( D \right)
\label{eq16:EDMMCollection}
\end{equation}
Hence, all agents obtain $P_{\mathcal{A}}(D)$ after sufficient iterations, which includes all of the neighbored EDMs. 
\begin{remark}
	%需要强调的是，对于分布式无向图的邻接矩阵备份，除了平均一致共识算法外，最大共识算法也是一种有效的实现方式。在这个过程中，代理仍旧以$P_{{\mathcal{A}}_i}(D)$为初始状态。在每个迭代步，代理会与邻居交换信息，并通过保留邻域内代理LEDMM相应位置的最大值，来更新自身的状态。这样，每个代理就获得$P{{\mathcal{A}}( D)$备份。事实上，这种实现方式不仅保证全体代理获得一致的$P{{\mathcal{A}}( D)$，还能以更快的速度达成共识。
	It is important to note that another efficient approach for EDM collection is the maximum consensus algorithm. Agents take $P_{{\mathcal{A}}_i}(D)$ as their initial status and update their status by retaining the maximum value of the corresponding position of the neighboring agent's LEDMM at each iteration step until the consensus is reached. In fact, this implementation not only ensures that all agents obtain a consistent $P_{\mathcal{A}}( D)$, but also achieves consensus at a faster rate.
\end{remark}

\subsubsection{Non-neighbored EDMs estimation}
\label{subsubsec:EDMMCompletionWithFixedRank}
Next, based on the network-wide consistent $P_{\mathcal{A}}(D)$, the agents independently perform low-rank matrix completion to estimate non-neighbored EDMs:
\begin{equation}
\mathop {\min}\limits_{\scriptstyle \tilde D \in {\mathbb{R}^{N \times N}}, rank ( {\tilde D} ) \leqslant s } f( {\tilde D} ) = \frac{1}{\lambda}\left\| {{P_{\cal A}}( {D - \tilde D} )} \right\|_F^2 + \left\| \text{diag}( \tilde D ) \right\|_F^2
\label{eq13:AimFunction}
\end{equation}
where ${{\left\| \cdot \right\|}_{\text{F}}}$ is the Frobenius norm; $\tilde{D}$ is the low-rank approximation of $D$; $\text{diag}(\tilde D)$ is the vector consisting of the main diagonal elements of $\tilde D$; and $\lambda$ is the regularization parameter. This objective function explores the matrix space of $rank(\tilde D) \leqslant s$ to determine $\tilde{D}$, in which $\left\| {{P_{\cal A}}( {D - \tilde D} )} \right\|_F^2$ guarantees that $\tilde D$ is an approximation of $D$ and $\text{diag}(\tilde D)$ is used to minimize the modulus of the main diagonal elements of $\tilde D$ as $Dismp(\boldsymbol{m}_i,\boldsymbol{m}_i) = 0$.

Searching directly in the matrix space of $rank(\tilde{D})\le s$ tends to yield $\tilde{D}$ with inaccurate rank, which reduces the precision of credibility. Hence, the optimization of Eq.(\ref{eq13:AimFunction}) is decomposed into two iterative subtasks: optimization on the fixed-rank smooth Riemannian manifold ${\mathcal{M}}_k$ of rank $k\le s$ and rank adjustment of the optimization result.

\subsubsection{Fixed-rank optimization}
\label{subsubsec:Fixed-rank optimization}

In fixed-rank manifold optimization, the objective function becomes:
\begin{equation}
\begin{aligned}
& \underset{\tilde{D}}{\mathop{\min }}\,f( {\tilde{D}} )=\frac{1}{\lambda}\left\| {{P}_{\mathcal{A}}}( D-\tilde{D} ) \right\|_{F}^{2}+\left\| diag( {\tilde{D}} ) \right\|_{F}^{2} \\ 
& s.t.\quad \tilde{D}\in {{\mathcal{M}}_{k}=\{ \tilde{D}\in {{\mathbb{R}}^{N\times N}}:rank( {\tilde{D}} )=k \}} 
\end{aligned}
\label{eq17:AimFunctionFixedRank}
\end{equation}

According to the Riemannian gradient descent~\cite{gao2021ariemannian}, Eq.(\ref{eq17:AimFunctionFixedRank}) can be addressed by iteratively updating $ \tilde{D}(t)$:
\begin{equation}
\tilde{D}( t+1) ={{\mathcal{R}}_{\tilde{D}(t)}}(h(t)Z(t))
%	\begin{aligned}
%	&\\
%	={{\mathcal{P}}_{{{\mathcal{M}}_{k}}}}( \tilde{D}( t)+h( t)Z( t))
%	&= {{\mathcal{P}}_{{{\mathcal{M}}_{k}}}}\left( \tilde{D}( t)-h( t)\left( U{{U}^{T}}\nabla f(\tilde{D}(t))V{{V}^{T}}+{{U}_{\bot }}U_{\bot }^{T}\nabla f( \tilde{D}( t))V{{V}^{T}} +U{{U}^{T}}\nabla f( \tilde{D}( t)){{V}_{\bot }}V_{\bot }^{T} \right)\right)
%	\end{aligned}
\label{eq20:ProjectionRetractionOperator}
\end{equation}
with:
\begin{equation}
\begin{aligned}
Z(t) 	&\triangleq -{{\operatorname{grad}}_{k}}f(\tilde{D}(t))\\
%	=-{{\mathcal{P}}_{{{\text{T}}_{\tilde{D}(t)}}{{\mathcal{M}}_{k}}}}( \nabla f( \tilde{D}(t)))\\ 
&=-\left( UU^T\nabla f(\tilde{D}(t))V(t)V(t)^T+{{U(t)}_{\bot }}U(t)_{\bot }^{T}\nabla f( \tilde{D}( t))V(t){{V(t)}^{T}}\right.\\&\left. +U(t){{U(t)}^{T}}\nabla f( \tilde{D}( t)){{V}(t)_{\bot }}V(t)_{\bot }^{T} \right)
\end{aligned}
\label{eq19:RiemanianGradient}
\end{equation}
\begin{equation}
\nabla f( \tilde{D}( t ))=\frac{2}{\lambda}P_{\mathcal{A}}( \tilde{D}(t)-D )\mathcal{A}+2I\odot \tilde{D}(t)
\label{eq18:EuclideanGradientOfAimFunction}
\end{equation}
where $\mathcal{R}$ is the restriction operator that guarantees $\tilde{D}(t+1) \in {{\mathcal{M}}_{k}}$~\cite{absil2012projectionlike,vandereycken2013lowrank}; $Z(t)$ is the negative Riemannian gradient direction~\cite{absil2007optimization}; $U(t)\in {{\mathbb{R}}^{N\times k}}$ and $V(t)\in {{\mathbb{R}}^{N\times k}}$ are the left and right singular matrices of $\tilde{D}(t)$, respectively; ${{U}(t)_{\bot }}\in {{\mathbb{R}}^{N\times (N-k)}}$ and ${{V}(t)_{\bot }}\in {{\mathbb{R}}^{N\times (N-k)}}$ are the basis of the orthogonal complementary spaces of $U(t)$ and $V(t)$, respectively; $\nabla f( \tilde{D}( t ) )$ is the Euclidean gradient of $f$ at $\tilde{D}(t)$. A classical retraction called partitioned eigenvalue decomposition is given here:
\begin{align}
\left[Q_u,R_u\right] &= \text{qr}\left(-h(t) (I- U(t)U(t)^T )\nabla f(\tilde{D(t)}) V(t)\right)\\
[Q_v,R_v] &= \text{qr}\left(-h(t)(I- V(t)V(t)^T ) \nabla f(\tilde{D(t)})^T U(t) \right)\\
\left[ {{U_{\cal R}},{\Sigma _{\cal R}},{V_{\cal R}}} \right] &\leftarrow {\rm{svd}}\left( {\left[ {\begin{array}{*{20}{c}}
		{\Sigma(t)  - h(t) {U(t)^T}\nabla f(\tilde D(t))V(t)}&{\;\;\;{\kern 1pt} R_v^T}\\
		{{R_u}}&{\;\;\;{\kern 1pt} {\bf{0}}}
		\end{array}} \right],k} \right)\\
{\tilde D}(t+1) &= [U(t) \quad Q_u]U_{\mathcal{R}}\Sigma_{\mathcal{R}}V_{\mathcal{R}}^T[V(t) \quad Q_v]^T
\end{align}
where $\text{qr}$ denotes the QR decomposition performed on the matrix and $\text{svd}$ denotes the singular value decomposition performed on the matrix. $U_{\cal R} \in {{\mathbb{R}}^{N\times k}}$, $\Sigma _{\cal R} \in {{\mathbb{R}}^{k\times k}}$, and $V_{\cal R} \in {{\mathbb{R}}^{N\times k}}$ are results of singular value decomposition. $\Sigma(t)$ is the diagonal matrix formed by the first $k$ singular values of $\tilde D(t)$. The optimization step length $h(t)=\gamma (t){{\delta }^{\zeta}}$ is determined by the non-monotonic Armijo line search method with Barzilai-Borwein (BB) step~\cite{iannazzo2017the}, which seeks the smallest non-negative integer $\zeta$ that satisfies:
\begin{equation}
f( \tilde D(t+1)) =  f({\cal R}_{\tilde D(t)}(h(t)Z(t))) \leqslant c(t) + \beta \gamma( t){\delta ^\zeta }\langle {{{\operatorname{grad} }_k}f( {\tilde D( t)}),Z( t)}\rangle
\label{eq21:NonArijmoBBStepLength}
\end{equation}
where $\beta, \delta \in (0,1)$ are the fixed step length and step length discount factors, respectively; and $c(t)$ is determined by: 
\begin{equation}
c(t+1) = \frac{\theta q(t)c(t) + f\left( {\tilde D\left( t + 1 \right)} \right)} {q(t+1)}
\label{eq22:NonArijmoSearchPara}
\end{equation}
with $\theta \in [0,1]$, $q(0)=1$, and $q(t+1)=\theta q(t)+1$. And the full decay coefficient ${\gamma} (t)$ in Eq.(\ref{eq21:NonArijmoBBStepLength}) is computed as:
\begin{equation}
\gamma(t) = \max ( {{\gamma _{\min }},\min ( {\bar{\gamma} (t),{\gamma _{\max }}} )} )
\end{equation}
where $\bar{\gamma} (t)$ is:
\begin{equation}
\renewcommand{\arraystretch}{1.5}
\bar{\gamma} \left( t \right) = 
\begin{cases}
\dfrac{{\left\langle {S\left( t \right),S\left( t \right)} \right\rangle }}{{\left| {\left\langle {S\left( t \right),K\left( t \right)} \right\rangle } \right|}} &,  \text{if }t\text{ is odd,} \\[10pt]
\dfrac{{\left| {\left\langle {S\left( t \right),K\left( t \right)} \right\rangle } \right|}}{{\left\langle {K\left( t \right),K\left( t \right)} \right\rangle }} &,  \text{if } t \text{ is even.}
\end{cases}
\label{eq23:NonArijmoSearchGama}
\end{equation}
with $S(t) = h(t-1){{\cal T}_{\tilde D(t-1) \to \tilde D(t)}}( {Z(t-1)})$ and $K(t) = {{\cal T}_{\tilde D(t-1) \to \tilde D(t)}}(Z(t-1))-Z(t)$. The $\cal T$ is the vector transport on ${\cal M}_k$, which transports $Z(t-1)$ to $Z(t)$. Let:
\begin{align}
\Sigma_{\cal T} &= U(t)^TU(t-1)U(t-1)^T\nabla f(\tilde D(t-1))V(t-1)V(t-1)^TV(t)\\
U_{\cal T} &= (I - U(t))U(t)^T(I - U(t-1)U(t-1)^T)\nabla f(\tilde D(t-1))V(t-1)V(t-1)^TV(t)\\
V_{\cal T} &= (I\!-\!V(t))V(t)^T(I - V(t-1)V(t-1)^T)\nabla f(\tilde D(t\!-\!1))^TU(t-1)U(t-1)^TU(t)\\
T_O^N &= U_{\cal T}\Sigma_{\cal T}V_{\cal T}^T
\label{eq27:VectorTransportOnFiold}
\end{align}
then:
\begin{align}
S(t) &= h(t-1) \frac{\langle \text{grad}_sf(\tilde D(t-1)) ,\text{grad}_sf(\tilde D(t-1))\rangle}{\langle T_O^N,T_O^N\rangle}T_O^N\\
K(t) &= \text{grad}_sf(\tilde D(t)) + T_O^N
\label{eq28:VectorTransportOnFiold}
\end{align}

%	It should be noted that when computing $S(t)$ and $K(t)$, $Z(t-1)$ is transported to the tangent space of $Z(t)$ with the vector transport operator ${{\cal T}_{X \to Y}}:{{\text{T}}_X}{{\cal M}_k} \to {{\text{T}}_Y}{{\cal M}_k},\;{ \kern 1pt} Z \mapsto {{\cal P}_{{{\text{T}}}_Y{{\cal M}_k}}}(Z)$. 

On the basis of the above formulation, a fixed rank optimization algorithm is given as follows to recover the EDM between non-neighboring agents' pieces of evidence:
\begin{algorithm}[!h]
	\caption{Fixed rank optimization.}
	\label{alg3:Fixed rank optimization}
	\begin{algorithmic}[1]
		\REQUIRE To-be-completed EDMM $\tilde D(t)$, the rank of EDMM $k$, the adjacence matrix $\cal A$, the maximum number of searches for non-negative integers $Iter_{\zeta}$, $\delta$, $\gamma_{\min}$, $\gamma_{\max}$, $\theta$, $c(t)$, $\gamma(t)$, $q(t)$.
		\ENSURE EDMM $\tilde D(t+1)$, $c(t+1)$, $\gamma(t+1)$, $q(t+1)$.%%output
		\STATE $\zeta \leftarrow 1$.
		\IF {$\gamma (t)$ is not given as a input}
		\STATE $\gamma (t) \leftarrow \dfrac{\langle P_{\cal A}(\text{grad}_s f( \tilde{D}(t))), P_{\cal A}(\tilde D(t)-D) + \text{diag}(\tilde D(t))\rangle} {||P_{\cal A}(\text{grad}_s f( \tilde{D}(t)))||_F^2}$
		\ENDIF
		\WHILE {$\zeta \leq Iter_{\zeta}$}%${\cal R}_{\tilde D}(-h\text{grad}_sf(\tilde D)) = {\cal P}_{{\cal M}_s} \left({\tilde D} + -h\text{grad}_sf(\tilde D) \right)$
		\STATE $\tilde{D}( t+1) \leftarrow {\mathcal{R}_{\tilde{D}(t)}}(h(t)Z(t))$
		\IF {Eq.(\ref{eq21:NonArijmoBBStepLength}) is satisfied}
		\STATE break.
		\ENDIF
		\STATE $h(t) \leftarrow \gamma(t)\delta^{\zeta} $
		\STATE $\zeta \leftarrow \zeta + 1$
		\ENDWHILE
		\STATE Calculate $\bar{\gamma}(t+1)$ with Eq.(\ref{eq23:NonArijmoSearchGama}) and update $\gamma(t+1)$, $c(t+1)$, and $q(t+1)$.
	\end{algorithmic}
\end{algorithm}
\subsubsection{Rank adjustment}
\label{subsubsec:Rank adjustment}

To initiate the optimization of Eq.(\ref{eq13:AimFunction}), an initial guess for $rank(\tilde{D})$ is considered. If the guessed rank is incorrect, then Eq.(\ref{eq17:AimFunctionFixedRank}) is not an equivalent form of Eq.(\ref{eq13:AimFunction}), leading to an increase in estimation error for credibility. Therefore, it becomes essential to adjust the $rank(\tilde{D})$ to ensure high precision in the estimation of non-neighbored EDMs. The following inequality is for judging if $rank(\tilde{D}) = k <s$ is too small~\cite{gao2021ariemannian}:
\begin{equation}
\| N_{s-k}( \tilde{D} ) \|_{\text{F}} >\epsilon \| G_k( \tilde{D} ) \|_{\text{F}}
\label{eq25:JudgementForRankTooSmall}
\end{equation}
where $\epsilon$ is a positive number, and $N_{s-k}( \tilde{D} )$ and $G_{k}( \tilde{D} )$ are projections of $-f( {\tilde{D}} )$ in the normal subspace $( T_{\tilde{D}}\mathcal{M}_k )_{\le s-k}^{\bot }$ and the tangent space $T_{\tilde{D}}\mathcal{M}_k$, respectively. If Eq.(\ref{eq25:JudgementForRankTooSmall}) holds, the following formula is used to obtain the matrix after rank increase:
\begin{equation}
\tilde{ \tilde {D}} 	= \begin{bmatrix}	U & W	\end{bmatrix}\begin{bmatrix}	\Sigma & 0 \\	0 & \alpha H	\end{bmatrix}\begin{bmatrix}	V & Y	\end{bmatrix}
\label{eq28:RankIncrease}
\end{equation}
with:
\begin{equation}
W = {U_ \bot }[ {{{\bar u}_1}, \ldots ,{{\bar u}_{\tilde l}}} ], H = \operatorname{diag} ( {{{\bar \sigma }_1}, \ldots ,{{\bar \sigma }_{\tilde l}}} ),
Y = {V_ \bot }[ {{{\bar v}_1}, \ldots,{{\bar v}_{\tilde l}}} ]
\label{eq27:DefinitionOfWHY}
\end{equation}
\begin{equation}
\alpha = - \frac{{\langle {( {{\cal A} + I} ) \odot WH{Y^T},( {{\cal A} + I} ) \odot ( {\tilde D - D} )} \rangle }}{{\| {( {{\cal A} + I} ) \odot WH{Y^T}} \|_{\text{F}}^2}}
\label{eq29:alpha}
\end{equation}
where $\Sigma \triangleq diag( {{\sigma }_{1}},\ldots ,{{\sigma }_{k}} )$ is the diagonal matrix composed of the eigenvalues of $\tilde{D} \in \mathcal{M}_k$, i.e., there is $\tilde{D}=U\Sigma {{V}^{T}}$; $\bar U \triangleq \left[ {{{\bar u}_1}, \ldots,{{\bar u}_{\mathchar'26\mkern-10mu r}}} \right]$, $\bar \Sigma \triangleq \operatorname{diag} \left( {{{\bar \sigma }_1}, \ldots,{{\bar \sigma }_{\mathchar'26\mkern-10mu r}}} \right)$, and $\bar V \triangleq \left[ {{{\bar v}_1}, \ldots,{{\bar v}_{\mathchar'26\mkern-10mu r}}} \right]$ are the SVD result of $- U_ \bot ^ \top \nabla f( {\tilde D} ){V_ \bot }$ with ${\mathchar'26\mkern-10mu r} = rank(U_ \bot U_ \bot ^ \top \nabla f( {\tilde D} ){V_ \bot }V_ \bot ^ \top)$; $\tilde{l}=\min ( l,r )$ is the value of rank increment, where $l \le s - k$ is the upper bound of rank-increasing for the current operation.

Let ${{\sigma }_{1}}\ge \cdots \ge {{\sigma }_{k}}>0$, then the condition for determining that $rank(\tilde{D})$ is too large is:
\begin{equation}
\max\left(\frac{(\sigma _i-\sigma_{i+1})}{\sigma_{i}}\right) >\Delta
\label{eq31:ConditionRankReduce}
\end{equation}
where $\Delta > 0$ is the threshold for rank reduction. If Eq.(\ref{eq31:ConditionRankReduce}) holds, the rank-reduced matrix is obtained via the singular value truncation. Let $\tilde r = {\arg \min }_i \{ \sigma _i  \geq \Delta \sigma_1\}  < k$ be the target value of $rank(\tilde{D})$. Taking the first $\tilde{r}$ columns of $U$ and $V$ as ${{U}_{{\tilde{r}}}}$ and ${{V}_{{\tilde{r}}}}$ and ${{\Sigma }_{{\tilde{r}}}}$ as the first $\tilde{r}$ rows and $\tilde{r}$ columns of $\Sigma $, it obtains the rank-reduced matrix $\tilde {D} ={{U}_{{\tilde{r}}}}{{\Sigma }_{{\tilde{r}}}}V_{{\tilde{r}}}^{T}\in {{\mathcal{M}}}_{{\tilde{r}}}$.
%Whether $rank(\tilde{D})$ is too large is determined based on the singular values of $\tilde{D}$~\cite{32_zhou_2016_a}. The magnitude of the matrix singular values represents the importance of their corresponding components in the matrix. The singular values of matrices that can be approximated by low-rank matrices tend to obey the power-law distribution, i.e., the sum of a few singular values has a large weight in the sum of all singular values. Therefore, retaining the larger singular values will maximize the retention of matrix information. The magnitude of the matrix singular values represents the importance of their corresponding components in the matrix. The singular values of matrices that can be approximated by low-rank matrices tend to obey the power-law distribution, i.e., the sum of a few singular values has a large weight in the sum of all singular values. Therefore, retaining the larger singular values will maximize the retention of matrix information. Given the threshold for rank reduction , it indicates that $rank(\tilde{D})$ is too large if there exist ${{\sigma }_{i}}<{{\sigma }_{1}}\Delta ,i=1,\cdots ,k$. When this happens, the singular value truncation is utilized to reduce $rank(\tilde{D})$. 

%In fact, it is impossible for agents to recover the raw evidence of others based on the completed EDMM, since the operations of Eq.(\ref{eq6:BetPDefinition}) and Eq.(\ref{eq3:DismPDefinition}) are irreversible.

\begin{remark}
	By iteratively performing fixed-rank optimization and rank adjustment, the agents will obtain a low-rank approximation $\tilde D$ of the EDMM, based on which the evidence credibility can be computed. In this paper, we assume that all agents follow the same optimization parameters and all use $P_{\mathcal{A}}(D)$ as the initial value of $\tilde D$, thus guaranteeing that the complementation of the EDMM is network-wide consistent.
\end{remark}

\subsection{Fusion network consensus}
\label{subsec: Fusion network consensus}
The globally consistent and element-complete EDMM provided in \ref{subsec: EDM network consensus} allows agents to effortlessly calculate the credibility of any given evidence locally, which not only facilitates the preprocessing of their own raw evidence but also introduces a new challenge, i.e., their neighbors are able to recover their raw evidence from the preprocessed one received by using the obvious inverse operator of Eq.(\ref{eq8:EvidenceDiscount}). It is therefore undesirable to share pre-processed evidence between agents directly.

%This subsection presents a two-stage privacy-preserving consensus strategy for distributed fusion of raw evidence, adhering to three principles: credible fusion, evidence security, and consistent result. The requirement for credible fusion involves computing credibility and preprocessing evidence. Evidence security askes that information sharing during fusion does not lead to the leakage of raw evidence. Consistent result necessitates that all agents converge to a consistent fusion result, and when the EDMM is accurately provided, the distributed fusion result aligns with CCEF.

%However, Eq.(\ref{eq30:EvidenceFusion_LAC}) fails to meet the requirements of credibility and security. On the one hand, there is the possibility of producing counterintuitive fusion result due to the lack of credibility assessment and evidence pre-processing. On the other hand, directly sharing the weight assignment of the agent's raw evidence as the initial state with its neighbors may lead to privacy leakage.

Obviously, Eq.(\ref{eq30:EvidenceFusion_LAC}) fails to meet the requirements of credibility and security for evidence fusion, as it directly shares the weight assignment of the agent's raw evidence as the initial state with its neighbors. To address this, a privacy-preserving term is introduced into Eq.(\ref{eq30:EvidenceFusion_LAC}):
\begin{equation}
\begin{aligned}
\boldsymbol{x}_i^{\boldsymbol{\omega}} \left( {t + 1} \right) &= \boldsymbol{x}_i^{\boldsymbol{\omega}} \left( t \right) + \sum\limits_{j \in {{\cal N}_i}}^N {{c_{ij}}\left( {\boldsymbol{x}_j^{\boldsymbol{\omega}} \left( t \right) - \boldsymbol{x}_i^{\boldsymbol{\omega}} \left( t \right)} \right)} + {\bf{u}}_i\left( {t + 1} \right) \\
\boldsymbol{x}_i^{\boldsymbol{\omega}} \left( {0} \right) &= {{\boldsymbol{\omega}'}_i} + {\bf{u}}_i\left( {0} \right)
\end{aligned}
\label{eq30:Proposed_EvidenceFusion_LAC}
\end{equation}
where ${\boldsymbol{\omega}'}_i$ is the weight assignment corresponding to the preprocessed evidence $\boldsymbol{m}'_i$; ${\bf{u}}_i \left( {t} \right)$ is the privacy-preserving %modification 
term that consists of a differential privacy term~\cite{schneider2015convergence} ${\bf{u}}_i^R \left( {t} \right)$ and a credibility compensation term ${\bf{u}}_i^{\boldsymbol{\omega}} \left( {t}\right)$:
\begin{equation}
{\bf{u}}_i\left( t \right) = {\bf{u}}_i^R \left( t \right) + {\bf{u}}_i^{\boldsymbol{\omega}} \left( t \right)
\label{eq31:privacy-preserving_modification_term}
\end{equation}

In Eq.(\ref{eq31:privacy-preserving_modification_term}), ${\bf{u}}_i \left( {t} \right)$ is used to mask the raw evidence, bringing uncertainty to the fusion result. To ensure consistent convergence, the following theorem is presented:
\begin{thm}
	\textbf{(Evidence fusion consensus)} If ${\bf{u}}_i^R \left( {t} \right)$ and ${\bf{u}}_i^{\boldsymbol{\omega}} \left( {t}\right)$ satisfy the following consensus condition:
	\begin{enumerate}[label=\arabic*)]
		\setlength{\itemsep}{0pt}
		\setlength{\parsep}{0pt}
		\setlength{\parskip}{0pt}
		\item {\textbf{privacy-preserving-free condition:}}
		\begin{enumerate}[label=\roman*.]
			\setlength{\itemsep}{0pt}
			\setlength{\parsep}{0pt}
			\setlength{\parskip}{0pt}
			\item ${\bf{u}}_i^R(t)$ is self-canceling, i.e., $\sum\nolimits_{t=0}^{{t_i}} {{\bf{u}}_i^R(t)} = \bf{0}$, where ${t_i} \leqslant {t_{\max }}$ is generated independently and randomly by Agent $i$ and is not less than the convergence time of EDMM with $t_{\max }$ is observed by all agents.
			\item ${\bf{u}}_i^R(t)$ is finite-time effect, i.e., $\sum\nolimits_{t = 0}^{t_i} {{{\left\| {{\bf{u}}_i^R(t)} \right\|}_2}} > 0$ and $\forall t > {t_i},{\bf{u}}_i^R(t) = {\bf{0}}$.
			%			\item ${\bf{u}}_i^R(t)$ has non-zero entries on $[0,t_i]$, i.e., .
		\end{enumerate}
		\item {\textbf{credibility compensation condition:}}The $u_{ij}^{\boldsymbol{\omega}} (t)$, the $j$-th component of ${\bf{u}}_i^{\boldsymbol{\omega}} \left( {t}\right)$, satisfies:
		\begin{equation}
		\begin{array}{l}
		u_{ij}^{\boldsymbol{\omega}} (t) = 
		\left\{ \begin{array}{ll}
		\sum\limits_{{A_k} \supseteq {A_j}} {{{(-1)}^{|A_k| - | A_j|}}\ln \frac{1 - Cred_{i}(t)(1 - Q_{i}(A_k))}{1 - Cred_{i}(t-1)(1-Q_{i}(A_k) )}}&, t > 0\\
		0{}&,t = 0
		\end{array} \right.
		\end{array}
		\label{eq31:u_omega_ij}
		\end{equation}
	\end{enumerate}
	then, the following conslusions hold:
	\begin{enumerate}[label=\arabic*)]
		\item All agents' states converge to $\sum\nolimits_{i = 1}^N {\omega _i^*} /N$ in which ${\omega _i^*}$ is the weight assignment of $Cred^*_i$-discount of $\boldsymbol{m}_i$ with $Cred^*_i$ being the credibility of $\boldsymbol{m}_i$ given the converged EDMM.
		\item The iterative fusion is independent of the differential privacy term.
		\item If the iterative EDMM converges to CCEF, then Eq.(\ref{eq31:privacy-preserving_modification_term}) converges to the fusion result of CCEF Eq.(\ref{eq9:DRforNEvidence}).
	\end{enumerate}
	\label{Thm1:TheoremOfFusionConsensus}
\end{thm}

\begin{pf}
	Here, the convergence of Eq.(\ref{eq30:Proposed_EvidenceFusion_LAC}) is proven using the $j$-th component of $\boldsymbol{x}_i^{\boldsymbol{\omega}}\!\left( t \right)$ as an example, since there is coupling among the components of $\boldsymbol{x}_i^{\boldsymbol{\omega}} \left( t \right)$. Define $\underline{\boldsymbol{x}}_{\cdot j}^{\boldsymbol{\omega}}(t)\!\!=\!\![x_{1j}^{\boldsymbol{\omega}}(t),\!x_{2j}^{\boldsymbol{\omega}}(t),\!\cdots\!,x_{Nj}^{\boldsymbol{\omega}}(t)]^T$, $\underline{\bf{u}}_{\cdot j}^{\boldsymbol{\omega}}(t)\!=\![u_{1j}^{\boldsymbol{\omega}}(t),\!u_{2j}^{\boldsymbol{\omega}}(t),\!\cdots\!,\!u_{Nj}^{\boldsymbol{\omega}}(t)]^T\!$, and $\underline{\bf{u}}_{\cdot j} ^{\boldsymbol{R}}(t)\!=\!{{[u_{1j}^{\boldsymbol{R}}(t),\!u_{2j}^{\boldsymbol{R}}(t),\!\cdots\!,\!u_{Nj}^{\boldsymbol{R}}(t)]}^{T}}$ as vectors consisting of the $j$-th components of the states, fusion modification terms, and differential privacy terms of all agents, respectively. Then, there is:
	\begin{equation}
	\begin{aligned}
	\underline{\boldsymbol{x}}^{\boldsymbol{\omega}}_{\cdot j}(t+1) &= \left[ {\begin{array}{*{20}{c}}
		{c_{11}}&{c_{12}}& \cdots &{c_{1N}}\\
		{c_{21}}&{c_{22}}& \cdots &{c_{2N}}\\
		\vdots & \vdots & \ddots & \vdots \\
		{c_{N1}}&{c_{N2}}& \cdots &{c_{NN}}
		\end{array}} 
	\right]
	\underline{\boldsymbol{x}}^{\boldsymbol{\omega}}_{\cdot j}(t) + \underline{\bf{u}}^{\boldsymbol{\omega}}_{\cdot j}(t+1) + \underline{\bf{u}}^R_{\cdot j}(t+1)\\
	& \buildrel \Delta \over = C\underline{\boldsymbol{x}}^{\boldsymbol{\omega}}_{\cdot j}(t) + \underline{\bf{u}}^{\boldsymbol{\omega}}_{\cdot j}(t+1) + \underline{\bf{u}}^R_{\cdot j}(t+1)
	\end{aligned}
	\end{equation}
	The limit of $\underline{\boldsymbol{x}}^{\boldsymbol{\omega}}_{\cdot j}(t)$ is expressed as the number of iterations tends to infinity:
	\begin{equation}
	\begin{aligned}
	\mathop {\lim }\limits_{t \to \infty } \underline{\boldsymbol{x}}^{\boldsymbol{\omega}}_{\cdot j}(t) 
	&= \mathop {\lim }\limits_{t \to \infty } C   \underline{\boldsymbol{x}}^{\boldsymbol{\omega}}_{\cdot j}(t-1) + \underline{\bf{u}}^{\boldsymbol{\omega}}_{\cdot j}(t) + \underline{\bf{u}}^R_{\cdot j}(t)\\ 
	&= \mathop {\lim }\limits_{t \to \infty } {C^2}\left(\underline{\boldsymbol{x}}^{\boldsymbol{\omega}}_{\cdot j}(t-2) + \underline{\bf{u}}^{\boldsymbol{\omega}}_{\cdot j}(t-1) +\underline{\bf{u}}^R_{\cdot j}(t - 1) \right) + \underline{\bf{u}}^{\boldsymbol{\omega}}_{\cdot j} (t) + \underline{\bf{u}}^R_{\cdot j}(t) \\ 
	&\vdots \\ 
	&= \mathop {\lim }\limits_{t \to \infty } \left( {C^t}\underline{\boldsymbol{x}}^{\boldsymbol{\omega}}_{\cdot j} \left( 0 \right) + \sum\limits_{\mathchar'26\mkern-10mu\lambda = 1}^t {{C^{t - \mathchar'26\mkern-10mu\lambda + 1}}\underline{\bf{u}}^{\boldsymbol{\omega}}_{\cdot j} \left( \mathchar'26\mkern-10mu\lambda \right)} + \sum\limits_{\mathchar'26\mkern-10mu\lambda = 1}^t {{C^{t - \mathchar'26\mkern-10mu\lambda + 1}}\underline{\bf{u}}^R_{\cdot j}\left( \mathchar'26\mkern-10mu\lambda \right)} \right)
	\end{aligned}
	\end{equation}
	Due to $\forall t>t_i$, $\underline{\bf{u}}^{\boldsymbol{\omega}}_{i}(t) = \underline{\bf{u}}^R_{i}(t) = \bf{0}$, therefore:
	\begin{equation}
	\begin{aligned}
	\mathop {\lim }\limits_{t \to \infty } \underline{\boldsymbol{x}}^{\boldsymbol{\omega}}_{\cdot j}(t) 
	&= \mathop {\lim }\limits_{t \to \infty } \left( {C^t}\underline{\boldsymbol{x}}^{\boldsymbol{\omega}}_{\cdot j} \left( 0 \right) + \sum\limits_{\mathchar'26\mkern-10mu\lambda = 1}^t {{C^{t - \mathchar'26\mkern-10mu\lambda + 1}}\underline{\bf{u}}^{\boldsymbol{\omega}}_{\cdot j} \left( \mathchar'26\mkern-10mu\lambda \right)} + \sum\limits_{\mathchar'26\mkern-10mu\lambda = 1}^t {{C^{t - \mathchar'26\mkern-10mu\lambda + 1}}\underline{\bf{u}}^R_{\cdot j}\left( \mathchar'26\mkern-10mu\lambda \right)} \right)\\
	&= \mathop {\lim }\limits_{t \to \infty } \left( {C^t}\underline{\boldsymbol{x}}^{\boldsymbol{\omega}}_{\cdot j} \left( 0 \right) + \sum\limits_{\mathchar'26\mkern-10mu\lambda = 1}^{t_i} {{C^{t - \mathchar'26\mkern-10mu\lambda + 1}} \left(\underline{\bf{u}}^{\boldsymbol{\omega}}_{\cdot j} \left( \mathchar'26\mkern-10mu\lambda \right) + \underline{\bf{u}}^{R}_{\cdot j} \left( \mathchar'26\mkern-10mu\lambda \right)\right)} +\!\!\! \sum\limits_{\mathchar'26\mkern-10mu\lambda = t_i + 1}^{t}\!\!\! {{C^{t - \mathchar'26\mkern-10mu\lambda + 1}}\!\! \left(\underline{\bf{u}}^{\boldsymbol{\omega}}_{\cdot j} \left( \mathchar'26\mkern-10mu\lambda \right) + \underline{\bf{u}}^{R}_{\cdot j} \left( \mathchar'26\mkern-10mu\lambda \right)\right)}\right)\\
	%	&= \mathop {\lim }\limits_{t \to \infty } \left( {C^t}\underline{\boldsymbol{x}}^{\boldsymbol{\omega}}_{\cdot j} \left( 0 \right) + \sum\limits_{\mathchar'26\mkern-10mu\lambda = 1}^{t_i} {{C^{t - \mathchar'26\mkern-10mu\lambda + 1}} \left(\underline{\bf{u}}^{\boldsymbol{\omega}}_{\cdot j} \left( \mathchar'26\mkern-10mu\lambda \right) + \underline{\bf{u}}^{R}_{\cdot j} \left( \mathchar'26\mkern-10mu\lambda \right)\right)} + \sum\limits_{\mathchar'26\mkern-10mu\lambda = t_i + 1}^{t} {{C^{t - \mathchar'26\mkern-10mu\lambda + 1}} \left(\bf{0} + \bf{0}\right)}\right)\\
	&= \mathop {\lim }\limits_{t \to \infty } \left( {C^t}\underline{\boldsymbol{\omega}}'_{\cdot j} + \sum\limits_{\mathchar'26\mkern-10mu\lambda = 0}^{t_i} {{C^{t - \mathchar'26\mkern-10mu\lambda + 1}}\underline{\bf{u}}^{\boldsymbol{\omega}}_{\cdot j} \left( \mathchar'26\mkern-10mu\lambda \right)} + \sum\limits_{\mathchar'26\mkern-10mu\lambda = 0}^{t_i} {{C^{t - \mathchar'26\mkern-10mu\lambda + 1}}\underline{\bf{u}}^R_{\cdot j}\left( \mathchar'26\mkern-10mu\lambda \right)} \right)
	\end{aligned}
	\end{equation}
	where ${\underline{\boldsymbol{\omega }}'_{\cdot j} = \left[ {\omega}'_1\left(A_j\right),{\omega}'_2\left(A_j\right), \cdots ,{\omega}'_N\left(A_j\right) \right]^T}$. According to Eqs.(\ref{eq31:CommonalityFunction2WeightAssignment})-(\ref{eq32:CommonalityFunction}), there is:
	\begin{equation}
	{\omega '_i}\left( {{A_j}} \right) = \sum\limits_{{A_k} \supseteq {A_j}} {{{\left( { - 1} \right)}^{\left| {{A_k}} \right| - \left| {{A_j}} \right|}}\ln \left( {1 - Cred_{i}\left( 0 \right)\left( {1 - {Q_i}\left( {{A_k}} \right)} \right)} \right)}
	\end{equation}
	As $C$ is a doubly stochastic matrix, it follows that:
	\begin{equation}
	\begin{aligned}
	\mathop {\lim }\limits_{t \to \infty } \underline{\boldsymbol{x}}^{\boldsymbol{\omega}}_{\cdot j} \left( t \right) 
	= &\mathop {\lim }\limits_{t \to \infty } {C^t}\underline{\boldsymbol{\omega }}'_{\cdot j}
	+ \mathop {\lim }\limits_{t \to \infty } \sum\limits_{\mathchar'26\mkern-10mu\lambda = 0}^{{t_i}} {{C^{t - \mathchar'26\mkern-10mu\lambda }}\underline{\bf{u}}^{\boldsymbol{\omega }}_{\cdot j} \left( \mathchar'26\mkern-10mu\lambda \right)} 
	+ \mathop {\lim }\limits_{t \to \infty } \sum\limits_{\mathchar'26\mkern-10mu\lambda = 0}^{{t_i}} {{C^{t - \mathchar'26\mkern-10mu\lambda }}\underline{\bf{u}}^R_{\cdot j} \left( \mathchar'26\mkern-10mu\lambda \right)} \\
	=& \frac{1}{N}\left( {\sum\limits_{i = 1}^N {{{\omega}'_i}\left(A_j\right)} + \sum\limits_{i = 1}^N {\sum\limits_{\mathchar'26\mkern-10mu\lambda = 0}^{{t_i}} {u_{ij}^\omega (\mathchar'26\mkern-10mu\lambda )} } + \sum\limits_{i = 1}^N {\sum\limits_{\mathchar'26\mkern-10mu\lambda = 0}^{{t_i}} {u_{ij}^R(\mathchar'26\mkern-10mu\lambda )} } } \right) \cdot {{\bf{1}}_N} \\
	=& \frac{1}{N}\sum\limits_{i = 1}^N {\left( {{{\omega '}_i}\left( {{A_j}} \right) + u_{ij}^\omega (0) + \sum\limits_{\mathchar'26\mkern-10mu\lambda = 1}^{{t_i}} {u_{ij}^\omega (\mathchar'26\mkern-10mu\lambda )} } \right)} \cdot {{\bf{1}}_N} \\
	= &\frac{1}{N}\sum\limits_{i = 1}^N \sum\limits_{{A_k} \supseteq {A_j}} {{{\left( { - 1} \right)}^{\left| {{A_k}} \right| - \left| {{A_j}} \right|}}\ln \left( {1 - Cred_{i}\left( 0 \right)\left( {1 - {Q_i}\left( {{A_k}} \right)} \right)} \right)} \cdot {{\bf{1}}_N} \\
	&+ \frac{1}{N}\sum\limits_{i = 1}^N \sum\limits_{\mathchar'26\mkern-10mu\lambda = 1}^{{t_i}} {\sum\limits_{{A_k} \supseteq {A_j}} {{{\left( { - 1} \right)}^{\left| {{A_k}} \right| - \left| {{A_j}} \right|}}\ln \frac{{1 - Cred_{i}\left( \mathchar'26\mkern-10mu\lambda \right)\left( {1 - {Q_i}\left( {{A_k}} \right)} \right)}}{{1 - Cred_{i}\left( {\mathchar'26\mkern-10mu\lambda - 1} \right)\left( {1 - {Q_i}\left( {{A_k}} \right)} \right)}}} }  \cdot {{\bf{1}}_N} \\
	%			&= \frac{1}{N}\sum\limits_{i = 1}^N {\left( {\sum\limits_{{A_k} \supseteq {A_j}} {{{\left( { - 1} \right)}^{\left| {{A_k}} \right| - \left| {{A_j}} \right|}}\ln \left( {1 - Cred_{i}\left( 0 \right)\left( {1 - {Q_i}\left( {{A_k}} \right)} \right)} \right)} 
	%					+ \sum\limits_{\mathchar'26\mkern-10mu\lambda = 1}^{{t_i}} {\sum\limits_{{A_k} \supseteq {A_j}} {{{\left( { - 1} \right)}^{\left| {{A_k}} \right| - \left| {{A_j}} \right|}}\ln \frac{{1 - Cred_{i}\left( \mathchar'26\mkern-10mu\lambda \right)\left( {1 - {Q_i}\left( {{A_k}} \right)} \right)}}{{1 - Cred_{i}\left( {\mathchar'26\mkern-10mu\lambda - 1} \right)\left( {1 - {Q_i}\left( {{A_k}} \right)} \right)}}} } } \right)} \cdot {{\bf{1}}_N} \\
	= &\frac{1}{N}\sum\limits_{i = 1}^N {\left( {\sum\limits_{{A_k} \supseteq {A_j}} {{{\left( { - 1} \right)}^{\left| {{A_k}} \right| - \left| {{A_j}} \right|}}\ln \left( {1 - Cred_{i}\left( {{t_i}} \right)\left( {1 - {Q_i}\left( {{A_k}} \right)} \right)} \right)} } \right)} \cdot {{\bf{1}}_N}
	\end{aligned}
	\end{equation}
	Therefore, the fusion result is free from $\bf{u}^R$. As the EDMM has converged at iteration step $t_i$, it follows that $Cred_i(t_i) = Cred^*_i$, and:
	\begin{equation}
	\begin{aligned}
	\mathop {\lim }\limits_{t \to \infty } \underline{\boldsymbol{x}}^{\boldsymbol{\omega}}_{\cdot j} \left( t \right) 
	&= \frac{1}{N}\sum\limits_{i = 1}^N {\left( {\sum\limits_{{A_k} \supseteq {A_j}} {{{\left( { - 1} \right)}^{\left| {{A_k}} \right| - \left| {{A_j}} \right|}}\ln \left( {1 - Cred_{i}\left( {{t_i}} \right)\left( {1 - {Q_i}\left( {{A_k}} \right)} \right)} \right)} } \right)} \cdot {{\bf{1}}_N} \\
	&= \frac{1}{N}\sum\limits_{i = 1}^N {\boldsymbol{\omega }_i^*\left( {{A_j}} \right)} \cdot {{\bf{1}}_N} \\
	\end{aligned}
	\label{eq38:ProveOfTheorem_1_Conclusion}
	\end{equation}
	The Eq.(\ref{eq38:ProveOfTheorem_1_Conclusion}) is valid for each component of $\boldsymbol{x}^{\boldsymbol{\omega }}_i$. Therefore, after a sufficient number of iterations, all agents' states will converge consistently to ${\sum\nolimits_{i=1}^{N}{\boldsymbol{\omega}_{i}^{*}}}/{N}$. When EDMM is accurately provided, the credibility computed by each agent is identical to that of centralized fusion. In this case, the fusion result $\sum\nolimits_{i=1}^{N}{\boldsymbol{\omega}_{i}^{*}}$ is precisely equivalent to that of CCEF.
	
	Hence, the Theovem \ref{Thm1:TheoremOfFusionConsensus} is proven.
\end{pf}

In fact, Theorem \ref{Thm1:TheoremOfFusionConsensus} not only describes the consistency of fusion but also ensures the fusion process is credible. The ${{\boldsymbol{\omega}'}_i}$ is determined by $Cred_i$ whose value is related to the execution order of EDMM completion optimization and distributed fusion. Depending on whether Eq.(\ref{eq13:AimFunction}) and Eq.(\ref{eq30:Proposed_EvidenceFusion_LAC}) are executed simultaneously, the execution order is classified into two modes: serial and parallel. The serial execution mode starts the iteration of Eq.(\ref{eq30:Proposed_EvidenceFusion_LAC}) until the EDMM output from Eq.(\ref{eq13:AimFunction}) completely converges, which makes Agent already know before Eq.(\ref{eq30:Proposed_EvidenceFusion_LAC}) is executed. Thus, ${\bf{u}}_i^{\boldsymbol{\omega}} \left( {t} \right) = {\bf{0}}$.

The parallel execution mode begins the iteration of Eq.(\ref{eq30:Proposed_EvidenceFusion_LAC}) while the optimization of Eq.(\ref{eq13:AimFunction}) is still ongoing. Indeed, once the neighbored EDMs are collected, the agents can already recover the non-neighboring EDMs. And although this recovery is not precise for some time at the beginning, it is sufficient to allow the agents to compute $Cred_i(t)$ based on the element-complete EDMM and to start the execution of Eq.(\ref{eq30:Proposed_EvidenceFusion_LAC}). Noting that $\boldsymbol{\omega}'_i(t)$ is the $Cred_i(t)$-discount of $\boldsymbol{m}$, one has accordingly $\boldsymbol{\omega}'_i=\boldsymbol{\omega}'_i(0)$. Before EDMM has fully converged, $Cred_i(t)$ will keep changing, which leads to the existence of a gap between $\boldsymbol{\omega}'_i(t)$ and $\boldsymbol{\omega}'_i(t-1)$, which is why the ${\bf{u}}_i^{\boldsymbol{\omega}}(t)$ is introduced. And as $\tilde{D}(t)$ tends to converge, the amplitude of $u_{ij}^{\boldsymbol{\omega}}(t)$ becomes progressively smaller. When fully converged, there is $u_{ij}^{\boldsymbol{\omega}}(t) = 0$. The difference between agents' states at this step is smaller than the initial value of the serial execution approach because Eq.(\ref{eq30:Proposed_EvidenceFusion_LAC}) has been executed for some time, which means that they will reach agreement in fewer iterative steps, i.e., in favor of lowering the PCEF's elapsed time.

Regarding security, the following theorem ensures the privacy of evidence for each agent under Eq.(\ref{eq30:Proposed_EvidenceFusion_LAC}).
\begin{thm}
	\textbf{(Privacy-preserving of evidence fusion)} Agent $i$'s raw evidence is not inferred by Agent $j \in {\cal N}_i$ if and only if $\left\{ {\cal N}_i,i \right\}\not \subset {{\cal N}_j}$.
	\label{Thm2:TheoremOfFusionSecurity}
\end{thm}
\begin{pf}
	As Agent $i$ shares its state at any iteration step with its neighbors, Agent $j$ is able to infer:
	\begin{equation}
	\begin{aligned}
	\boldsymbol{x}_i^{\boldsymbol{\omega}} (t + 1) - (1 -\!\! \sum\limits_{l \in {{\cal N}_i}} {{c_{il}}} )\boldsymbol{x}_i^{\boldsymbol{\omega }} (t) -\!\!\! \sum\limits_{l \in {{\cal N}_i} \cap {{\cal N}_j}}\!\!\! {{c_{il}}\boldsymbol{x}_l^{\boldsymbol{\omega }} (t)} 
	&= \!\!\!\!\!\sum\limits_{l \in {{\cal N}_i},l \notin {{\cal N}_j}}\!\!\!\!\! {{c_{il}}\boldsymbol{x}_l^{\boldsymbol{\omega }} (t)} + {\bf{u}}_i^R(t + 1) + {\bf{u}}_i^{\boldsymbol{\omega }} (t + 1) \\
	\boldsymbol{x}_i^{\boldsymbol{\omega}} (0) &= {\boldsymbol{\omega}'_i} + {\bf{u}}_i^R(0) + {\bf{u}}_i^{\boldsymbol{\omega}} (0)
	\end{aligned}
	\label{eq39:ProveOfTheorem_2_Conclusion}
	\end{equation}
	
	The left side of Eq.(\ref{eq39:ProveOfTheorem_2_Conclusion}) represents the known quantity for Agent $j$, while the right side represents the unknown. The recovery of other agents' raw evidence by Agent $j$ is contingent upon eliminating the differential privacy term. According to Theorem \ref{Thm1:TheoremOfFusionConsensus}, the way to eliminate the differential privacy term is to accumulate Eq.(\ref{eq39:ProveOfTheorem_2_Conclusion}):
	\begin{equation}
	\begin{aligned}
	&\sum\limits_{t = 0}^{{t_i} + 1} {\boldsymbol{x}_i^{\boldsymbol{\omega}} (t)} - \sum\limits_{t = 0}^{{t_i}} {(1 - \sum\limits_{l \in {{\cal N}_i}} {{c_{il}}} )\boldsymbol{x}_i^{\boldsymbol{\omega}} (t)} + \sum\limits_{t = 0}^{{t_i}} {\sum\limits_{l \in {{\cal N}_i} \cap {{\cal N}_j}} {{c_{il}}\boldsymbol{x}_l^{\boldsymbol{\omega}} (t)} } \\
	&= \sum\limits_{t = 0}^{{t_i}} {\sum\limits_{l \in {{\cal N}_i},l \notin {{\cal N}_j}} {{c_{il}}\boldsymbol{x}_l^{\boldsymbol{\omega}} (t)} } + {\boldsymbol{\omega }'_i} + \sum\limits_{t = 0}^{{t_i}} {{\bf{u}}_i^{\boldsymbol{\omega}} (t)} + \sum\limits_{t = 0}^{{t_i}} {{\bf{u}}_i^R(t)} \\
	&= \sum\limits_{t = 0}^{{t_i}} {\sum\limits_{l \in {{\cal N}_i},l \notin {{\cal N}_j}} {{c_{il}}\boldsymbol{x}_l^{\boldsymbol{\omega}} (t)} } + {\boldsymbol{\omega }'_i} + \sum\limits_{t = 0}^{{t_i}} {{\bf{u}}_i^{\boldsymbol{\omega}} (t)} \\ 
	&= \sum\limits_{t = 0}^{{t_i}} {\sum\limits_{l \in {{\cal N}_i},l \notin {{\cal N}_j}} {{c_{il}}\boldsymbol{x}_l^{\boldsymbol{\omega}} (t)} } + {\boldsymbol{\omega}} _i^* 
	\end{aligned}
	%	\label{eq39:ProveOfTheorem_2_Conclusion}
	\end{equation}
	It is clearly that if $\left\{ {\cal N}_i,i \right\}\subset {{\cal N}_j}$, then:
	
	\begin{equation}
	\begin{aligned}
	\sum\limits_{t = 0}^{{t_i} + 1} {\boldsymbol{x}_i^{\boldsymbol{\omega}} (t)} - \sum\limits_{t = 0}^{{t_i}} {(1 - \sum\limits_{l \in {{\cal N}_i}} {{c_{il}}} )\boldsymbol{x}_i^{\boldsymbol{\omega}} (t)} + \sum\limits_{t = 0}^{{t_i}} {\sum\limits_{l \in {{\cal N}_i} \cap {{\cal N}_j}} {{c_{il}}\boldsymbol{x}_l^{\boldsymbol{\omega}} (t)} } 
	&= {\boldsymbol{\omega}} _i^* 
	\end{aligned}
	%	\label{eq39:ProveOfTheorem_2_Conclusion}
	\end{equation}
	Therefore, Agent $j$ recovers $\boldsymbol{m}_i$ from ${\boldsymbol{\omega}} _i^* $, as it is able to obtain $Cred_i^*$ according to the EDMM given in \ref{subsec: EDM network consensus}.
	
	If there is Agent $l \not \in \mathcal{N}_j$ in $\mathcal{N}_i$, then Agent $j$ cannot get $\boldsymbol{m}_i$ because it does not know anything about $\boldsymbol{x}_l^{\boldsymbol{\omega}}(t)$.
	
	Thus, the Theorem \ref{Thm2:TheoremOfFusionSecurity} is proven.
	
\end{pf}

According to Theorem \ref{Thm2:TheoremOfFusionSecurity}, we can conclude as follows:
\begin{proposition}
	The privacy of all raw evidence will be preserved during PCEF if all agents satisfy Theorem \ref{Thm2:TheoremOfFusionSecurity}.
\end{proposition}
\begin{proposition}
	In a fully connected network, the privacy of raw evidence cannot be preserved.
\end{proposition}

\begin{algorithm}[!h]
	\caption{PCEF.}
	\label{alg2:PCEF}
	\begin{algorithmic}[1]
		\REQUIRE evidence $\{ {\boldsymbol{m}_1},{\boldsymbol{m}_2}, \cdots ,{\boldsymbol{m}_N}\} $, local adjacency matrices $\{ {{\cal A}_1},{{\cal A}_2}, \cdots ,{{\cal A}_N}\} $, a guess for the rank $k$, a guess for the rank maximum $s$, and an upper bound on a single rank increase $l$, the threshold for rank reduction $\Delta >0$, the parameter for rank increase $\epsilon>0$, the maximum iterative steps for rank adjustments $Iter_{RA}$, the maximum iterative steps for LAC $Iter_{consen}$, a time for all agents to agree $t_{max} < Iter_{consen}$, the maximum iterative steps for rank unchanged $Iter_{RUc}$.
		\ENSURE Fusion result ${\boldsymbol{m}_{\oplus }}$.%%output
		\STATE Agents construct their LEDMM $\bar D_i = P_{{\cal A}_i}(D)$.
		\STATE Localize $P_{\cal A}(D)$ and $\cal A$ to each agent with a iterative number of $Iter_{consen}$.
		\STATE $\left[U,\Sigma,V\right] \leftarrow \text{svd}(P_{\cal A}(D) , k)$.
		\STATE $\tilde r \leftarrow \max \left\{\left(\sigma _i - \sigma_{i+1}\right) \mathord{\left/\right.} {\sigma_i}\right\}$.
		\IF {$k > \tilde r$}
		\STATE $k \leftarrow \tilde r$.
		\STATE $[U,\Sigma,V] \leftarrow \text{svd}(P_{\cal A}(D) , k)$.
		\STATE $\tilde D(0)  \leftarrow U\Sigma V^T$.
		\ENDIF
		\FOR {each Agent $i \in [1,N]$, in parallel}
		\STATE Generate a random number $t_i$ that satisfies $Iter_{RA} < t_i < Iter_{consen}$.
		\STATE $t \leftarrow 1$.
		\WHILE{$t \leq [1,Iter_{consen} + Iter_{RA}]$}
		\IF{$t\leqslant Iter_{RA}$ \textbf{and} $k$ has not been constant for the past $Iter_{RUc}$ iteration steps}
		\STATE Get $\tilde D(t)$ from Algorithm \ref{alg3:Fixed rank optimization} with $\tilde D(t-1)$ as the initial value.
		\IF{$\tilde r = {\arg \min }_i \{ \sigma _i  \geq \Delta \sigma_1\}  < k$ \textbf{and} $\max \left\{\left(\sigma _i - \sigma_{i+1}\right) \mathord{\left/\right.} {\sigma_i}\right\} > \Delta$ }
		\STATE $k \leftarrow \tilde r$.
		\STATE Construct $U_{\tilde r}$, $\Sigma_{\tilde r}$, and $V_{\tilde r}$.
		\STATE $\tilde D(t) \leftarrow U_{\tilde r}\Sigma_{\tilde r}V_{\tilde r}$.
		\ELSIF{$k<s$  \textbf{and}  ${\| {{N_{s - k}}( {{{\tilde D}(t)}} )} \|_{\rm{F}}} >  \epsilon{\| {{G_k}( {{{\tilde D}(t)}})} \|_{\rm{F}}}$}
		\STATE Construct $W$, $H$, $Y$ and calculate $\alpha$.
		\STATE $\tilde D(t) \leftarrow \tilde D(t) + \alpha WH{Y^T}$.
		\STATE Rearrange $\tilde D(t)$ in descending order of singular values.
		\STATE $k \leftarrow k+\tilde l$.
		\ENDIF
		\STATE $t_{RA} \leftarrow t$.
		\ELSE 
		\IF {$t-t_{RA} > Iter_{consen}$}
		\STATE break.
		\ENDIF
		\ENDIF
		\STATE Calculate $Cred_i(t)$ according to $\tilde D(t)$ and update agent state $\boldsymbol{x}^{\boldsymbol{\omega }}_i(t)$ with Eq.(\ref{eq30:Proposed_EvidenceFusion_LAC}).
		\ENDWHILE	
		\ENDFOR
		\STATE Agents convert $N\boldsymbol{\omega}_i(t)$ to ${\boldsymbol{m}_ \oplus }$ by Eq.(\ref{eq33:WeightAssignment2MassFunction}).
	\end{algorithmic}
\end{algorithm}

In fact, Eq.(\ref{eq30:Proposed_EvidenceFusion_LAC}) provides a two-stage privacy-preserving strategy for distributed evidence fusion. In the first stage, agents protect their individual information by introducing self-canceling random noise into the system. In the second stage, the differential privacy term is set to zero to ensure the convergence of agent states after the information of agents is well mixed. While the introduction of the differential privacy term does not affect the fusion result, its magnitude influences the time-consuming of consensus. Considering that the credibility compensation term decreases as the EDMM converges, it is suggested to progressively reduce the magnitude of the differential privacy term to facilitate a quicker consensus among agents.

The pseudo-code for PCEF is given as Alg.\ref{alg2:PCEF}, in which the EDMM completion parameters are set to be the same for all agents. Thus, the EDM collection is executed only once. This algorithm adopts several maximum numbers of iteration parameters to govern termination of iterative processes because it includes such consensus or optimization as EDM collection, EDMM completion, and distributed credible evidence fusion, which makes it difficult to provide generic iterative termination conditions for all distributed networks.
\begin{remark}
	%虽然本文研究无向图下的分布式证据可信融合，但PCEF很容易被应用在具有生成树的有向图当EDM neighbor consensus阶段不强调保护原始证据的隐私。
	Although PCEF is designed for undirected graphs, it is also suitable for directed graphs with a spanning tree. In this case, information is shared unidirectionally among some agents, which means that the EDM between these agents cannot be obtained in the EDM neighbor consensus.
\end{remark}

\subsection{Computational complexity}
\label{subsec: Computational complexity}
The computational complexity of PCEF is analyzed below: 

\begin{enumerate}[label=\arabic*)]
	\setlength{\itemsep}{0pt}
	\setlength{\parsep}{0pt}
	\setlength{\parskip}{0pt}
	\item \textbf{EDM neighbor consensus}: Agent $i$ needs to compute $Dismp$ with its $|{\cal N}_i|$ neighbors, which operation involves inner product computation, privacy-preserving two-party dot-product protocols, and Millionaire Problem Solving algorithms. The computational complexity for the vector inner product computation is lower to negligible than the latter two. According to the literature \cite{wang2009design,liu2022design}, the computational complexities of the privacy-preserving two-party dot-product protocol and the Millionaire's Problem Solving algorithm are $O(5(2^n-1)+1)$ and $O(6n+4)$, respectively. Therefore, the computational complexity of EDM neighbor consensus is up to $O(\max\{|{\cal N}_i|\}(5\cdot 2^{n}+6n))$.
	\item \textbf{EDM network consensus}: The computational complexity of this module can be analyzed in the following parts:
	\begin{itemize}
		\setlength{\itemsep}{0pt}
		\setlength{\parsep}{0pt}
		\setlength{\parskip}{0pt}
		\item $P_{\cal A}(D)$ localization: At each iterative step, the computational complexity for agent $i$ to update its own state is $O(|{\cal N}_i|N^2)$. The worst computational complexity for $Iter_{consen}$ iterations is $O(Iter_{consen}\max\{|{\cal N}_i|\}N^2)$.
		\item Fixed-rank optimization: The computational complexity of both the matrix addition and the Hadamard product involved in computing $f(\tilde D(t))$ is $O(N^2)$. The computational complexity of the singular value decomposition of $\tilde D(t)$ is $O(kN^2)$. The computational complexity of the matrix multiplication included in the Frobenius paradigm computation of the gradient $\text{grad}_sf(\tilde D(t))$ and updating $\gamma$, $p$ and $q$ is $O(N^3)$. In the worst case, the computational complexity of searching for the optimal BB step is $O(Iter_{\zeta}N^3)$. Therefore, the computational complexity of this part is $O((Iter_{\zeta}+2)N^3+(k+1)N^2)$.
		\item Rank increase/reduction operation: The computational complexity of the decrease rank operation is $O(kN^2)$ because it mainly involves the multiplication operation of matrices. The complexity of the rank-increase operation is higher than the rank-decrease operation. The computational complexity for singular value decomposition of $-U_{\bot}^T\nabla f(\tilde D)V_{\bot}$ is $O((s - k)N^2)$. The operations with higher computational complexity for updating $\tilde D$ are singular value decomposition, QR decomposition, and matrix multiplication, all of which are $O(N^3)$. Therefore, the computational complexity of this part is $O(3N^3+sN^2)$.
	\end{itemize}
	In the worst case, the fixed-rank complement optimization and the increase/decrease rank operation are performed $Iter_{RA}$ times. Therefore, the computational complexity of EDM network consensus is $O(Iter_{RA}((Iter_{\zeta}+5)N^3+(Iter_{consen}\max{|{\cal N}_i|}+k+s+1)N^2))$.
	\item \textbf{Evidence fusion consensus}: This consensus involves the computation of $\bf{u}$ and the updating of agent state $\textbf{x}_i^{\boldsymbol{\omega}}$. It is executed $Iter_{consen}+Iter_{RA}$ times in the worst case, so the computational complexity is $O((Iter_{consen}+Iter_{RA})(\max{|{\cal N}_i|}+(2^n-1)) (2^n - 2))$.
	\item \textbf{Transformation of $\boldsymbol{\omega}$ to $\boldsymbol{m}_\oplus$}: It should execute DR $2^n-3$ times to transform $\boldsymbol{\omega}$ into $\boldsymbol{m}_\oplus$, each transformation with a complexity of $O(2^{2n})$. Consequently, the overall computational complexity of obtaining the global fusion result is $O((2^n-3)\cdot2^{2n})$
\end{enumerate}
In summary, the total computational complexity of PCEF is $O((5\cdot 2^{n}+6n)\max{|{\cal N}i|} + (Iter{\zeta}+5)N^3 + (Iter_{conses}\max{|{\cal N}_i|}+k+s+1)N^2 + (Iter_{ consen}+Iter_{RA})(\max{|{\cal N}_i|}+(2^n-1)) (2^n - 2) + (2^n-3)\cdot2^{2n})$.
\section{Simulation}
\label{sec:Simulation}
%本节首先借助一个数字仿真实验测试PCEF在秩估计、可信度估计和抵抗高冲突反常上的性能。然后列举一个无人集群协同的野生保护动物种群监测任务，对比PCEF与现有的两种分布式证据融合方法RANSAC-Based和COF-Based的融合性能。
In this section, the performance of PCEF on rank estimation, credibility estimation, and resistance to counterintuitive knowledge is first measured based on a numerical simulation experiment. Afterwards, the proposed method is compared with two existing distributed evidence fusion methods (i.e., RANSAC-Based and COF-Based methods) in a simulation that takes distributed unmanned swarm radar signal binning as the background.
\subsection{Verification for the approximation of PCEF to CCEF}
\label{subsec: Verification for the approximation of PCEF to CCEF}
%本仿真旨在测试PCEF的性能，包括验证其对EDMM秩的估计是否准确、证据可信度计算结果是否高精度逼近CCEF、融合高冲突证据是否不产生反常识结果等。考虑一个设计五个类别的分类任务。如Fig.\ref{fig3:5categoryPDF},所示，5个类别的概率密度曲线均为方差为1的正态分布。对应于类别 $\{a\}$, $\{b\}$, $\{c\}$, $\{d\}$, $\{e\}$, 概率密度函数的平均值分别为-2, -1, 0, 1, 2。
%假定一个连接密度为0.4的强连接无向图中的100个代理独立地观察一个类别为$\{a\}$的目标$T$。这些代理观测目标 $T$ 并将观测转化为一条定义在$\Omega = \{a, b, c, d, e\}$的证据。其中，代理$1\sim 90$的观测正常，服从均值为 -2、方差为 1 的正态分布，而代理$91\sim 100$的观测异常，服从均值为 1或2、方差为 1 的正态分布。Fig.\ref{fig3:5categoryPDF}中的圆形标记和星号标记分别代表90个正常观测值和10个异常观测值。在实验中，BKNN~\cite{liu2013anew}被作为基础分类器，以将代理们的观测值转化为$Omega$ 上的mass function，分别记为$\boldsymbol{m}_1 \sim \boldsymbol{m}_{100}$。BKNN的参数设置为：$\gamma _{t_a} = 2$、$\gamma _{t_r} = 5$、$K=20$ 和 $N_s = 100$。为了保证代理们能协同地融合证据，PCEF参数被设置如Tab.\ref{tab2:PCEFPara}。

\begin{figure}[!h]
	\centering
	\includegraphics[width=0.7\linewidth]{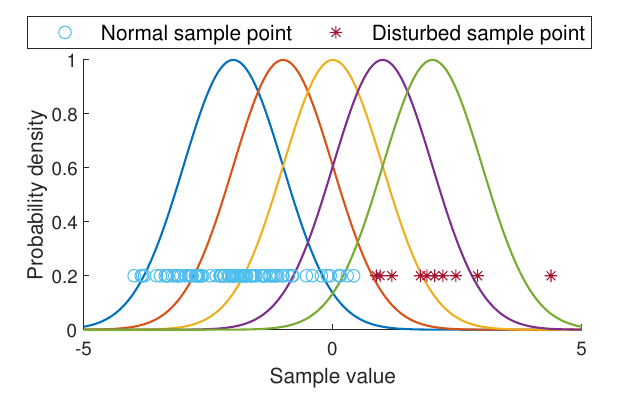}
	\caption{The probability density functions of 5 categories.}
	\label{fig3:5categoryPDF}
\end{figure}

This simulation aims to test the performance of PCEF, including if the estimated EDMM rank is accurate, if the computed evidence credibility approximates CCEF, and if the fusion of highly conflicting evidence produces counterintuitive results. Consider a classification task involving five categories whose probability density curves are all normally distributed with variance 1, as shown in Fig.\ref{fig3:5categoryPDF}. Corresponding to the categories $\{a\}$, $\{b\}$, $\{c\}$, $\{d\}$, $\{e\}$, the probability density functions have mean values of -2, -1, 0, 1, 2, respectively.

\begin{table}[!h]
	\setlength\tabcolsep{1mm
	\small
	\caption{Parameters involved in Alg.\ref{alg2:PCEF}}
	\label{tab2:PCEFPara}
	\centering
	%\scriptsize
	\begin{tabular}{cccccccccccccccc}
		\hline %[2pt]设置线宽
		Para. Name & $\Delta$ & $\epsilon$ & $l$ & $Iter_{RA}$&$k$&$s$&$\beta$&$\theta$&$\gamma_{\min}$&$\gamma_{\max}$&$Iter_{RUc}$&$Iter_{consen}$& $\lambda$&$\delta$&$Iter_{\zeta}$\\\hline %[2pt]
		Value & 0.1 & 10 & 1 & 200&10&36&1e-4&0.9&1e-15&1e15&20&100&2&0.1&5\\
		\hline %[2pt]
	\end{tabular}}
\end{table}

In a strongly connected undirected graph with a connection density of 0.4, 100 agents independently observe a target $T$ of category $\{a\}$. The observations of Agent $1\sim 90$ are normal and follow a normal distribution with mean -2 and variance 1, while the observations of Agent $91\sim 100$ are abnormal and follow a normal distribution with mean 1 or 2 and variance 1. The circular markers and asterisk markers in Fig.\ref{fig3:5categoryPDF} are for 90 normal observations and 10 abnormal observations, respectively. To transform these observations into pieces of evidence, denoted as $\boldsymbol{m}_1 \sim \boldsymbol{m}_{100}$, BKNN~\cite{liu2013anew} is adopted as the basic classifier in the simulation. The parameters of the BKNN are set as $\gamma _{t_a} = 2$, $\gamma _{t_r} = 5$, $K=20$, and $N_s = 100$. See \cite{liu2013anew} for the exact meaning of these parameters. In addition, to guarantee that the agents are able to fuse evidence collaboratively, the parameters of PCEF are set as in Tab.\ref{tab2:PCEFPara}. 

\begin{figure*}[!h]
	\centering
	\begin{subfigure}{0.45\linewidth}
		\centering
		\includegraphics[width=1\linewidth]{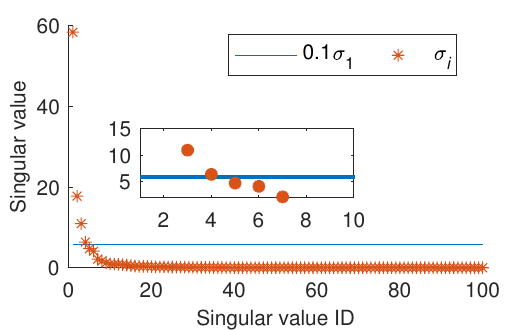}
		\caption{Singular values of $D$ ( Listed from largest to smallest).}
		\label{fig4:SingularValues}
	\end{subfigure}
	\begin{subfigure}{0.45\linewidth}
		\centering
		\includegraphics[width=1\linewidth]{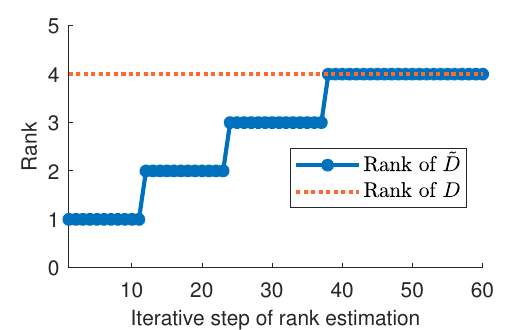}
		\caption{The rank of $\tilde D$.}
		\label{fig5:RankEstimation}
	\end{subfigure}
	\begin{subfigure}{0.45\linewidth}
		\centering
		\includegraphics[width=1\linewidth]{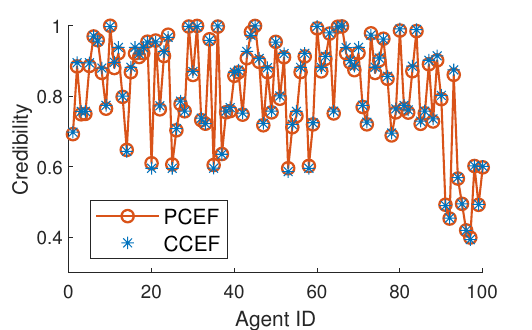}
		\caption{Credibility estimation results.}
		\label{fig6:Credibility}
	\end{subfigure}
	\begin{subfigure}{0.45\linewidth}
		\centering
		\includegraphics[width=1\linewidth]{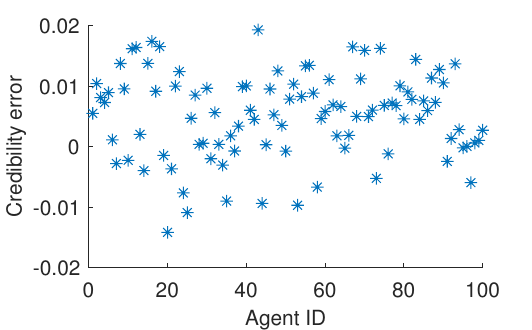}
		\caption{Estimation error of credibility.}
		\label{fig7:EstimationError}
	\end{subfigure}		
	\caption{Simulation results of PCEF}
\end{figure*}

In low-rank matrix completion, it is appropriate to regard the ordinate of the last singular value that is not less than the $\Delta$ time of the largest singular value as the rank of the matrix. Here, the EDMM with 100 pieces of evidence is constructed using $Dismp$ as the EDM. The singular values of this EDMM are computed and arranged in descending order, as shown in Fig.\ref{fig4:SingularValues}. It can be seen that there are four singular values of $D$ greater than $\Delta \sigma_1$ when $\Delta$ is set to 0.1 according to Tab.\ref{tab2:PCEFPara}, i.e., $rank(D)=4$.

%准确的秩估计能平衡Eq.\ref{eq17:AimFunctionFixedRank}优化效率和精度。如Fig.\ref{fig5:RankEstimation}所示，EDMM的低秩逼近矩阵的秩被记录。可以看到，$rank(\tilde D)$在60个迭代步中逐步被修正到并在连续的$Iter_{RUc} = 20}$个迭代步中保持不变。也就是说，PCEF对$rank(D)$的估计是准确的。
An accurate rank estimation balances the efficiency and accuracy of the optimization of Eq.(\ref{eq17:AimFunctionFixedRank}). The rank of the low-rank approximation matrix of the EDMM is recorded as shown in Fig.\ref{fig5:RankEstimation}. It is observed that $rank(\tilde D)$ is progressively corrected over 60 iteration steps and remains constant over successive $Iter_{RUc} = 20$ iteration steps. Namely, the estimation of $rank(D)$ by PCEF is accurate.

%EDMM补全的目的是评估待融合证据的可信度。由于PCEF是CCEF的思想在隐私保护分布式系统的应用，因此本仿真以CCEF给出的证据可信度为基准，评评价可信度的估计精度。各代理证据的可信度和可信度误差分别见图\ref{fig6:Credibility}和图\ref{fig7:EstimationError}。在图\ref{fig6:Credibility}中，CCEF给出的证据可信度用蓝星表示，PCEF估计的可信度用红圈表示。在图\ref{fig7:EstimationError}中，可信度误差用蓝星标记表示。可以看出，所有待融合证据的可信度估计误差都落在$[-0.02,0.02]$区间内。这表明本文提出的基于低秩矩阵补全技术的分布式系统证据可信度估计策略是可行的，具有较高的精度。
The purpose of completing the EDMM is to assess the credibility of the to-be-fused evidence. Since PCEF is applying the idea process of CCEF to privacy-preserving distributed systems, this simulation evaluates the accuracy of credibility calculated with PCEF by utilizing the evidence credibility given by CCEF as the benchmark. The credibility and credibility errors of each agent's evidence are shown in Fig.\ref{fig6:Credibility} and Fig.\ref{fig7:EstimationError}, respectively. In Fig.\ref{fig6:Credibility}, the evidence credibilities given by CCEF are marked by blue stars, and the credibility estimated by PCEF is indicated by red circles. In Fig.\ref{fig7:EstimationError}, the credibility error is shown as the blue star marker. It is seen that the credibility estimation errors of all pieces of to-be-fused evidence fall in the interval $[-0.02,0.02]$. This suggests that the proposed strategy for estimating the evidence credibility based on the low-rank matrix completion technique in a distributed system is feasible and highly accurate.

\begin{table}[!h]
	\small
	\renewcommand\arraystretch{1.5}
	\caption{Fusion results with three method: DR, CCEF, and PCEF.}
	\label{tab3:FusionResult}
	\setlength{\tabcolsep}{5pt}
	\centering
	\begin{tabular}{cccc}
		\hline
		Fusion method & $\{a\}$ & $\{b\}$ & $\{a,b\}$ \\ \hline
		DR  & $\cdot$&1&$\cdot$ \\
		CCEF & 0.9526 &0.0441 &0.0033\\
		PCEF & 0.9586 &0.0385 &0.0029\\ \hline
	\end{tabular}
\end{table}
The PCEF is compared with the CCEF and DR to verify that it avoids counterintuitive results when fusing high-conflict pieces of evidence. There are high conflicts among the 100 pieces of evidence since $\boldsymbol{m}_{91} \sim \boldsymbol{m}_{100}$ are set as disturbed pieces of evidence. The fusion results for the three cases are shown in Tab.\ref{tab3:FusionResult}. It shows that the fusion result of DR assigns all the confidence to $\{b\}$ due to the high conflict between pieces of evidence, which deviates seriously from the true value $\{a\}$. While the PCEF and CCEF assign the highest confidence to $\{a\}$, with 0.9586 and 0.9526, respectively, which indicates that the proposed method solves the problems of distributed evidence credibility assessment and high-conflict evidence fusion. In terms of confidence, PCEF gives fusion results that are very close to those of CCEF. It can be said that PCEF is a successful promotion of CCEF in privacy-preserving distributed systems. Moreover, it is the credibility error that makes the PCEF assign a higher support of 0.9586 to $\{a\}$ than the CCEF.

\subsection{Comparison experiment}
\label{subsec:ComparisonExperiment}
\begin{figure}[!h]
	\centering
	\includegraphics[width=0.5\linewidth]{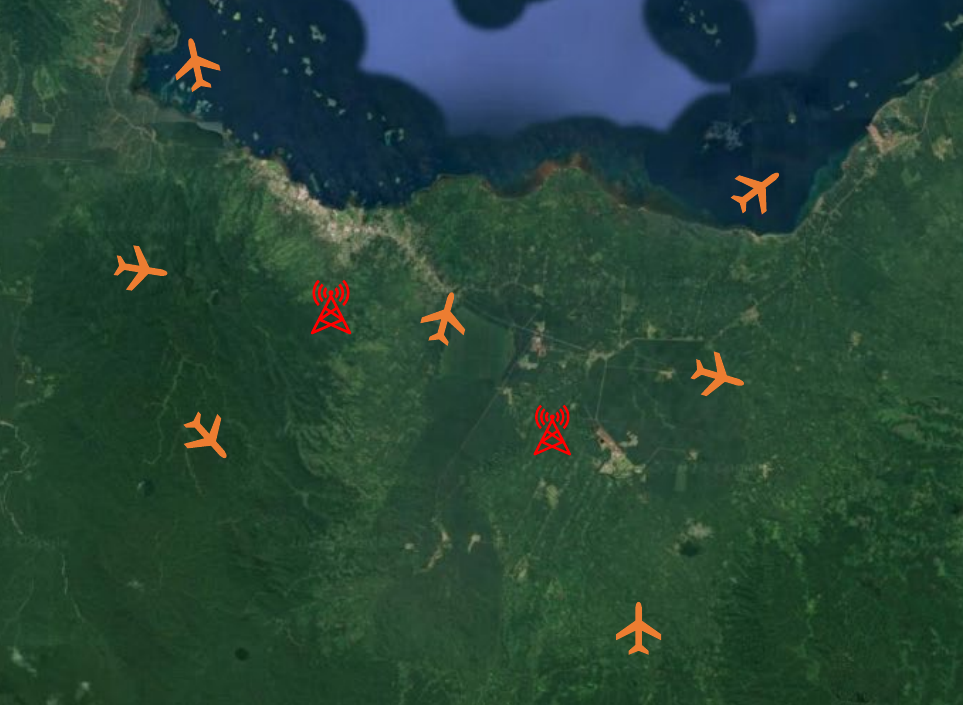}
	\caption{Collaborative battlefield reconnaissance using unmanned aerial vehicle swarm.}
	\label{fig8:UAVSwarmReconnaissance}
\end{figure}
%受复杂电磁环境影响和侦察探测能力限制，单架无人机往往不能满足电子战要求。近年来，基于分布式网络架构的无人飞行器蜂群已经被应用于大范围战场侦察和目标监测\cite{wan2022uav}。该集群不仅具有规模大、单机成本低、难以被探测设备发现等特点，还能通过搭载不同的探测设备实现信息互补。这种互补不仅体现在不同功能探测设备的多视图冗余，还体现在同功能设备的高低搭配。考虑到高精度设备往往被优先打击，保护平台信息安全就显得至关重要，这也是为什么本文强调保护代理原始证据的隐私。当然，同样的隐私要求还体现在协同地野生动物种群密度监测（防止非法盗猎）、污染物水平检测（防止污染倾倒）等应用中。如图6所示，协同地多UAV雷达信号分选被作为应用背景在本文以对比PCEF和现有两种分布式证据融合方法，RANSAC-Based和COF-Based，的性能。
%在考虑的场景中，一个由100架无人机组成的集群在广阔的战场范围内协同筛选雷达信号。这些信号可能来自五种不同的发射器，标记为$\Omega = \{a, b, c, d, e\}$。无人机首先独立进行信号筛选，然后通过自组织网络协议进行通信，协同实现集体决策。这一决策过程特别强调集群对相同单脉冲信号的识别和筛选。由于本仿真的重点不在于无人机的独立信号分选任务，因此它被假设所有无人机已经完成了独立信号筛选工作。
%四种方法CCEF、PCEF、RANSAC-Based、COF-Based被应用在该分选任务，它们的参数设置如下：
%\begin{itemize}
%	\item CCEF：与 \ref{subsec: Verification for the approximation of PCEF to CCEF}一样，已经被广泛验证的CCEF被作为基准。它遵循Eqs.(\ref{{eq3:DismPDefinition})-(\ref\ref{eq10:DRCentralized})的流程。
%	\item PCEF：遵循算法2的流程以及如\ref{subsec: Verification for the approximation of PCEF to CCEF}的参数设置。
%	\item RANSAC-Based：根据\cite{denux2021distributed}，RANSAC-Based的参数被设置为：the size of each random subsample $\nu = 5$, success probability $p_{suc} = 0.9999$, inlier probability $p_{in} = 0.9$, conflict threshold $\tau = 0.5$ that is suggest to be value in the range $[0.4,0.6]$.
%	\item COF-Based：根据\cite{zhao2023information}，COF-Based的参数设置为：the COF threshold $\tau_{COF} = 0.5$ and the distance threshold $\tau_{dist} = 0.7$ which are suggested with value in the ranges $[0.4,0.9]$ and $[0.4,0.7]$.
%\end{itemize}
Subject to the influence of the complex electromagnetic environment and the limitations of reconnaissance and detection capabilities, a single unmanned aerial vehicle (UAV) often fails to meet the requirements of electronic warfare. In recent years, UAV swarms based on distributed network technology have been applied to large-scale battlefield reconnaissance and target detection \cite{wan2022uav}. The swarm not only has the characteristics of large scale, low cost of a single aircraft, and being difficult to discover by detection equipment, but also provides complementary information by carrying different kinds of equipment. This complementarity is not only reflected in the multi-view information redundancy brought by hetero-functional devices but also in the high-low matching of same-functional devices. Considering that a high-precision device is often preferentially struck, it is crucial to protect the security of the platform's information, which is also why this paper emphasizes the privacy protection of the agent's raw evidence. Of course, the same privacy preservation requirements also appear in applications such as wildlife population density monitoring (to prevent illegal poaching) and pollutant level detection (to prevent pollution dumping).

As shown in Fig.\ref{fig8:UAVSwarmReconnaissance}, collaborative multi-UAV radar signal sorting~\cite{zhao2023information} is utilized as the background for application in this paper to compare the performance of PCEF and two existing distributed evidence fusion methods, RANSAC-Based and COF-Based. In the set scenario, a swarm of 100 UAVs collaborate to sort radar signals over a large-scale battlefield. These signals may come from five different transmitters labeled $\Omega = \{a, b, c, d, e\}$. The UAVs first perform signal sorting independently and then communicate via ad hoc networks to achieve collective decision-making. The UAVs first perform signal sorting independently and then communicate over an ad hoc network to collaborate on collective decision-making, which places special focus on the sorting of identical single-pulse signals. It is assumed that all UAVs have completed independent signal sorting, which is not the focus of this simulation.

Four methods, CCEF, PCEF, RANSAC-Based, and COF-Based, are applied to the sorting task, with their parameters set as follows:
\begin{itemize}
	\setlength{\itemsep}{0pt}
	\setlength{\parsep}{0pt}
	\setlength{\parskip}{0pt}
	\item CCEF: The CCEF, which has been widely validated, is used as the baseline in the same way as \ref{subsec: Verification for the approximation of PCEF to CCEF}. It follows the process Eqs.(\ref{eq3:DismPDefinition})-(\ref{eq10:DRCentralized}).
	\item PCEF: Follow the flow of Algorithm 2 and the parameter settings as in \ref{subsec: Verification for the approximation of PCEF to CCEF}.
	\item RANSAC-Based: According to \cite{denux2021distributed}, the parameters of RANSAC-Based are set to: the size of each random subsample $\nu = 5$, success probability $p_{suc} = 0.9999$, inlier probability $p_{in} = 0.9$, conflict threshold $\tau = 0.5$ that is suggested to be value in the range $[0.4,0.6]$.
	\item COF-Based: According to \cite{zhao2023information}, the parameters of COF-Based are set as follows: the COF threshold $\tau_{COF} = 0.5$ and the distance threshold $\tau_{dist} = 0.7 $ which are suggested with  values in the ranges $[0.4,0.9]$ and $[0.4,0.7]$.
\end{itemize}

\begin{sidewaysfigure}[htp]
	\centering
	\includegraphics[width=0.97\linewidth]{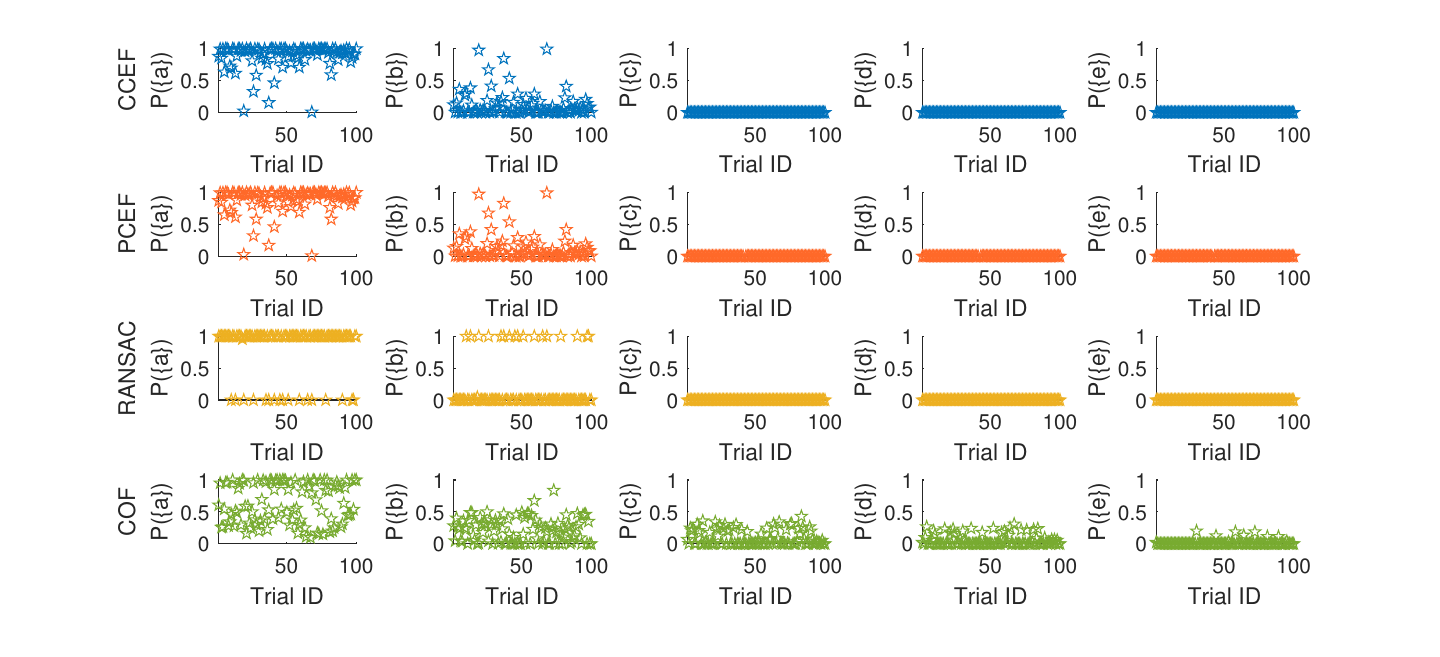}
	\caption{The Pignistic probabilities of the sorting results given by the four methods in 100 Monte Carlo trials.}
	\label{figPCEF_7:Comparison}
\end{sidewaysfigure}

\begin{sidewaysfigure}[htp]
	\centering
	\includegraphics[width=1\linewidth]{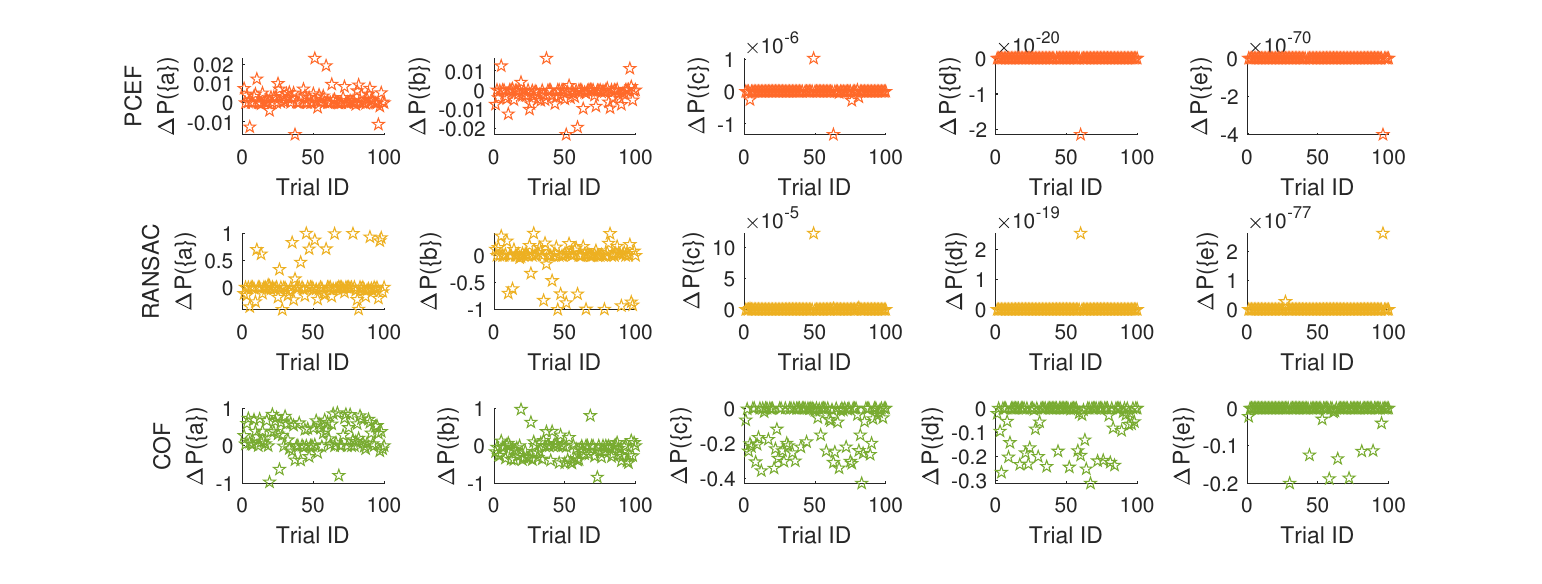}
	\caption{Error of three methods in 100 Monte Carlo trials.}
	\label{figPCEF_8:ClassificationAccuracy}
\end{sidewaysfigure}
%如\ref{subsec: Verification for the approximation of PCEF to CCEF}，UAVs构成的通信网络保持0.4的连接密度。为了避免单次测试带来偶然性结果，100次蒙特卡洛试验（Monte Carlo trials）被执行。Fig.\ref{figPCEF_7:Comparison}展示了四种方法的融合结果的Pignistic概率。可以看到，CCEF、PCEF和RANSAC-Based方法几乎不分配任何置信给类别$\{c\}$、$\{d\}$和$\{e\}$，而COF-Based方法则在多数试验中或多或少支持了这些类别。可能的原因是：在网络连接密度较高时，COF-Based方法基于邻域证据融合策略生成的初始融合中心很容易是反常的，这影响后续的投票结果。虽然同样基于异常检测思想辨识outliers证据，RANSAC-Based从全体证据中随机地采样在一定程度上规避了局部融合的反常。相比之下，PCEF评价证据的可信度在区间[0,1]中，并将不可信的信息折扣到FOD的全集以降低冲突。它避免了在inliers和outliers间的硬划分，更符合干扰情况下信源的随机特性。这也是为什么CEF在集中式融合中被广泛地验证是高度可靠的。PCEF、RANSAC-Based和COF-Based的Pignistic概率还被与CCEF的作差，如Fig.\ref{figPCEF_8:ClassificationAccuracy}。可以看到，基于高精度的可信度估计，PCEF获得与CCEF非常贴近的置信指派，误差不大于0.02，而RANSAC-Based和COF-Based方法则存在大量大差异结果。

As in \ref{subsec: Verification for the approximation of PCEF to CCEF}, the communication network constituted by the UAVs maintains a connection density of 0.4. Given that the result of a single trial may be by chance, 100 Monte Carlo trials are executed. The Pignistic probabilities of the fusion results obtained by the four methods are shown in Fig.\ref{figPCEF_7:Comparison}. It is clear that the CCEF, PCEF, and RANSAC-Based methods assign confidence to the categories $\{c\}$, $\{d\}$, and $\{e\}$ almost to be zero, whereas the COF-Based method supports these categories more or less in most of the trials. A possible reason for this phenomenon is that the initial fusion centers generated by the COF-Based method with the neighborhood evidence fusion strategy are easily anomalous when the network connection density is high, which affects the subsequent voting. Also based on anomaly detection to identify disturbed evidence, RANSAC-Based samples randomly from all pieces of evidence to circumvent local fusion anomalies to some extent. In contrast, PCEF evaluates the evidence credibility in the interval [0,1] and adds untrustworthy information into $m(\Omega)$ to reduce conflicts, which avoids hard division of evidence into inliers and outliers and is more in line with the stochastic nature of information in the presence of interference. It is also the reason why CEF is widely verified to be highly reliable in centralized fusion. Taking the Pignistic probability of CCEF as a baseline, the errors of PCEF, RANSAC-Based, and COF-Based are shown in Fig.\ref{figPCEF_8:ClassificationAccuracy}. It is observed that based on high precision credibility estimation, PCEF obtains very close confidence assignments to CCEF with errors no larger than 0.02, while RANSAC-Based and COF-Based methods have a large number of large discrepancy results. This suggests that PCEF is a more accurate extension of traditional centralized evidence fusion to distributed systems than RANSAC-Based method and COF-Based method.

\begin{figure}[!h]
	\centering
	\includegraphics[width=1\linewidth]{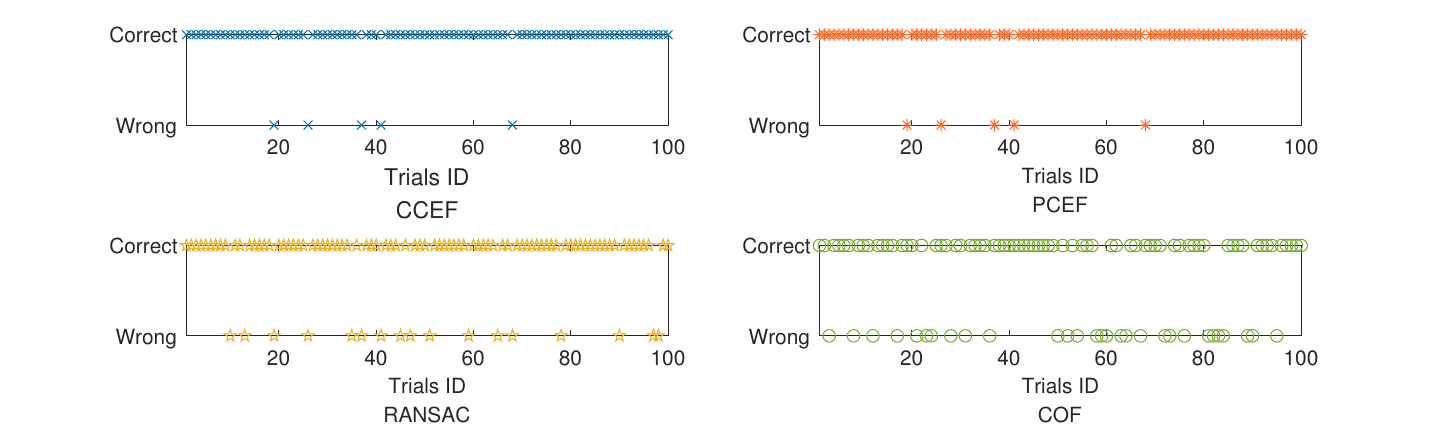}
	\caption{The correctness of the sorting decisions given by the four methods.}
	\label{figPCEF_9:SortingDecision}
\end{figure}

\begin{figure}[!h]
	\centering
	\includegraphics[width=0.8\linewidth]{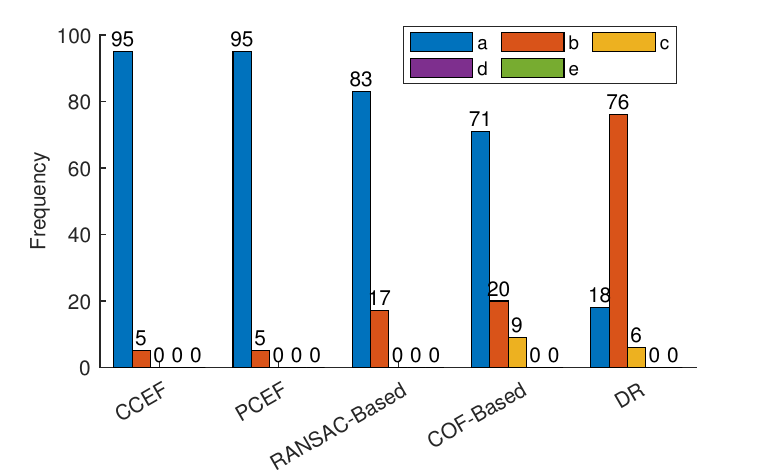}
	\caption{The frequencies of the sorting decision results given by the four methods and DR in 100 trials.}
	\label{fig10:ClassificationAccuracy}
\end{figure}
\begin{figure}[!h]
	\centering
	\includegraphics[width=0.8\linewidth]{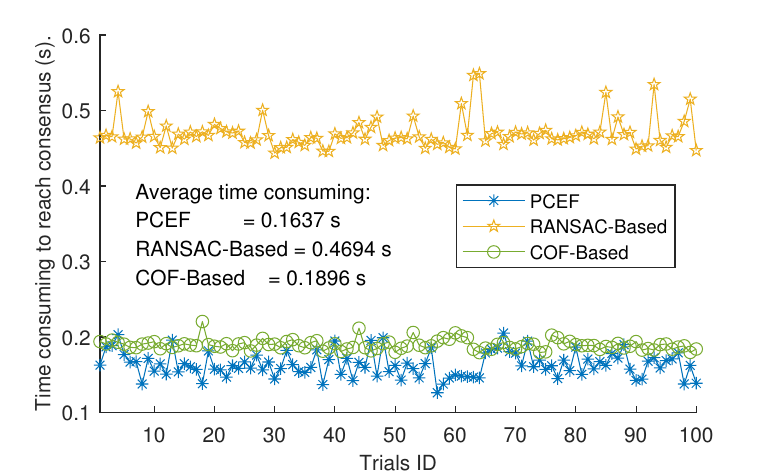}
	\caption{Time-consuming of PCEF, RANSAC-Based, and COF-Based methods.}
	\label{fig11:TimeConsuming}
\end{figure}

%\begin{figure*}[!h]
%	\centering
%	\begin{subfigure}{0.49\linewidth}
%		\centering
%		\includegraphics[width=1\linewidth]{ClassificationAccuracy.pdf}
%		\caption{The frequencies of the sorting decision results given by the four methods and DR in 100 trials.}
%		\label{fig10:ClassificationAccuracy}
%	\end{subfigure}
%	\begin{subfigure}{0.49\linewidth}
%		\centering
%		\includegraphics[width=1\linewidth]{TimeConsuming.pdf}
%		\caption{Time-consuming of PCEF, RANSAC-Based, and COF-Based methods.}
%		\label{fig11:TimeConsuming}
%	\end{subfigure}
%	\caption{Statistics for frequency of sorting decision results and time consuming of three distributed fusion methods.}
%\end{figure*}
%融合结果的Pignistic概率和最大Pignistic概率决策规则被采用完成最终的雷达信号分选决策。四种方法在100次试验中给出的决策如Fig.\ref{figPCEF_9:SortingDecision}所示。可以看到，PCEF比RANSAC_Based和COF-Based有更少的错误决策。Fig.\ref{fig10:ClassificationAccuracy}展示了CCEF、PCEF、RANSAC-Based、COF-Based和DR融合方法下的分选决策结果的频数统计。可以看到，三种分布式融合方法都明显改善了DR产生反常识融合结果的问题。其中，PCEF正确分选雷达信号的频数最高，且与CCEF保持一致，这说明基于矩阵补全技术估计证据可信度的策略是有效的。
The maximum Pignistic probability decision rule is employed to complete the final radar signal sorting decision. The decisions given by the four methods in 100 trials are shown in Fig.\ref{figPCEF_9:SortingDecision}. It can be seen that PCEF has fewer wrong decisions than RANSAC-Based and COF-Based. The Fig.\ref{fig10:ClassificationAccuracy} shows the frequency statistics of the sorting decision results under the CCEF, PCEF, RANSAC-Based, COF-Based, and DR fusion methods. It is observed that all three distributed fusion methods significantly improve the problem of DR producing counterintuitive results. In particular, the frequency counts of correctly sorting radar signals under PCEF are the highest and consistent with CCEF, which indicates that the strategy of estimating the evidence credibility based on the matrix-completion technique is effective.

%在本次实验中，我们对三种分布式融合方法的运行时间进行了测试。仿真实验是在安装有AMD 4800H处理器的计算机上使用MATLAB 2020b版本进行的，旨在测量每种融合方法的单个无人机执行时间。图9展示了每种方法在100次独立试验中的运行时间分布情况。结果显示，PCEF算法以平均0.1637秒的运行时间脱颖而出，展现出最快的处理速度。相比之下，RANSAC-Based和COF-Based方法的平均运行时间分别为0.4694秒和0.1896秒。
In this experiment, the runtimes of three distributed fusion methods are tested. The simulation experiments are performed using MATLAB 2020b on a computer with an AMD 4800H CPU and measure the individual UAV execution time for each fusion method. The time-consuming of each method over 100 independent trials is illustrated in Fig.\ref{fig11:TimeConsuming}, which shows that the PCEF algorithm stands out with an average time-consuming of 0.1637 seconds, demonstrating the fastest processing speed. In comparison, the RANSAC-Based and COF-Based methods have average time-consumings of 0.4694 seconds and 0.1896 seconds, respectively. Overall, PCEF achieves approximately a 12\% improvement in decision accuracy while consuming less time compared to existing methods.
\section{Conclusion}
This paper focuses on the distributed computation of evidence credibility and the privacy preservation of agents' raw evidence and proposes a distributed evidence credibility fusion method called PCEF that is applied to collective decision-making. It includes a three-layer consensus mechanism to overcome the problems of preference leakage and false positive/negative rates that exist in available distributed evidence fusion and is proved to be equivalent to CCEF when credibility is accurately given. In EDM neighbor consensus, precise computation of EDM between adjacent agents without revealing the raw evidence is accomplished by transforming it into two two-party secure computation subtasks. In EDM network consensus, the EDMM elements known by all agents are localized to each agent, and further, all missing elements of the EDMM are estimated with a rank-adaptive matrix completion technology. By doing so, credibility is estimated on the premise of privacy preservation. Leveraging the estimated plausibility, agents are instructed to credibly fuse evidence relying on the LAC, where active perturbations are added to protect privacy. Two experiments were conducted to validate the effectiveness of PCEF, whose results illustrate that PCEF successfully approximates CCEF and outperforms the existing methods both in terms of time-consuming and fusion accuracy.

In this paper, the calculation of neighbored EDMs is designed based on $Dismp$. In the future, we will explore the privacy computation methods for other candidate EDMs. In addition, the author will also introduce blockchain technology into the PCEF in the future to resist fusion failures caused by malicious attacks, which are very common in distributed swarm applications.

%\section*{CRediT authorship contribution statement}
%\textbf{Chaoxiong MA}: Conceptualization, Methodology, Software, Visualization, Data Curation, Writing - Original Draft. 
%\textbf{Yan LIANG}: Supervision, Funding Acquisition, Writing - Review \& Editing.
%
%\section*{Declaration of Competing Interest}
%The authors declare that they have no known competing financial interests or personal relationships that could have appeared to influence the work reported in this paper.

\section*{Acknowledgements}
This work is supported by the National Natural Science Foundation of China under Grant 61873205. It is also supported by the Innovation Foundation for Doctor Dissertation of Northwestern Polytechnical University through the Grand CX2023063.

%The authors would like to thank the anonymous reviewers for their detailed comments which helped to improve the quality of the paper.

\bibliography{Reference}
\end{document}